\documentclass[11pt]{article}

\usepackage[preprint]{acl}

\usepackage{times}
\usepackage{latexsym}

\usepackage[T1]{fontenc}
\usepackage[utf8]{inputenc}

\usepackage{microtype}

\usepackage{inconsolata}

\usepackage{graphicx}
\usepackage{url}
\usepackage{booktabs}
\usepackage{amsmath}        
\usepackage{amssymb}
\usepackage{amsfonts}
\usepackage{algorithm,algpseudocode}
\usepackage{graphicx}
\usepackage{enumitem}
\usepackage{multirow}
\usepackage{makecell}
\usepackage[normalem]{ulem}
\usepackage{caption}
\usepackage{subcaption}
\usepackage{xcolor}
\usepackage{tcolorbox}
\tcbuselibrary{skins}
\usepackage{hyperref}
\definecolor{linkblue}{HTML}{000099}
\hypersetup{
    colorlinks=true,
    allcolors=linkblue,
    pdfborder={0 0 0}
}

\definecolor{bsdoption}{HTML}{DCEAF5}
\definecolor{bsdreference}{HTML}{F7E3DB}
\definecolor{bsdpreprocessing}{HTML}{DDEDE6}
\definecolor{bsdambiguity}{HTML}{FFF0CD}

\newtcolorbox{bsdcase}[2]{
    enhanced,
    width=\linewidth,
    colback=black!1.5,
    colframe=black!40,
    boxrule=0.4pt,
    arc=1.5mm,
    boxsep=0pt,
    left=4pt,
    right=4pt,
    top=3pt,
    bottom=3pt,
    before skip=4pt,
    after skip=0pt,
    fontupper=\fontsize{9}{10}\selectfont,
    before upper={
        \setlength{\parindent}{0pt}
        \setlength{\parskip}{0pt}
        \textbf{#1}\hfill
        {\scriptsize\color{black!55}\texttt{#2}}
        \par\vspace{1pt}
    }
}

\newcommand{\bsdcasefield}[2]{%
    \noindent\textbf{#1:} #2\par\vspace{1pt}%
}

\newcommand{\bsdcasefooter}[5]{%
    \noindent
    {\footnotesize\color{black!65}%
        \textbf{Category:}\,
        \colorbox{#1}{\strut\textit{#2}}%
        \hfill
        $\widehat o_q^{\mathrm{ref}}=#5$\par
        $\mathbf{k}_q=(#3)$;\;
        $D_q=#4$%
    }\par
}

\title{Are Benchmarks Reliable? Toward Structural Diagnosis via Item-Level Capability Boundaries}

\author{%
Haiquan Hu$^{1,2}$ \quad
Yuzhu Liang$^2$ \quad
Weicheng Tang$^1$ \quad
Yanzeng Li$^1$ \\
\textbf{Yao Shi}$^1$ \quad
\textbf{Li Xu}$^3$ \quad
\textbf{Tian Wang}$^{2,}$\thanks{Corresponding Author}\\
$^1$Faculty of Arts and Sciences, Beijing Normal University\\
$^2$Institute of Artificial Intelligence and Future Networks, Beijing Normal University\\
$^3$College of Computer and Cyber Security, Fujian Normal University\\
\texttt{huhq99@yeah.net} \quad \texttt{cs\_tianwang@163.com}
}

\begin{document}
\maketitle
\begin{abstract}
Evaluating large language models (LLMs) relies heavily on benchmark scores, yet aggregate metrics can obscure whether individual items serve their intended measurement purpose. BSDProbe provides an item-level framework for benchmark structural diagnosis that jointly characterizes correctness transitions and their supporting response evidence. Repeated responses along an empirically ordered model axis yield complementary correctness and normalized entropy trajectories, from which four-dimensional profiles of boundary position, boundary width, boundary-signal validity, and order consistency are constructed for benchmark-level aggregation. Experiments on six benchmarks identify varying proportions of potentially anomalous items. With Qwen3 as both the selection and reference axis, BSDProbe achieves a 17.9\% mean relative improvement in normalized discrimination ratio over the best baseline per benchmark and matches or exceeds all baselines in ranking agreement with the full benchmark.
\end{abstract}

\section{Introduction}
\begin{figure}[ht]
  \centering
  \includegraphics[width=\columnwidth]{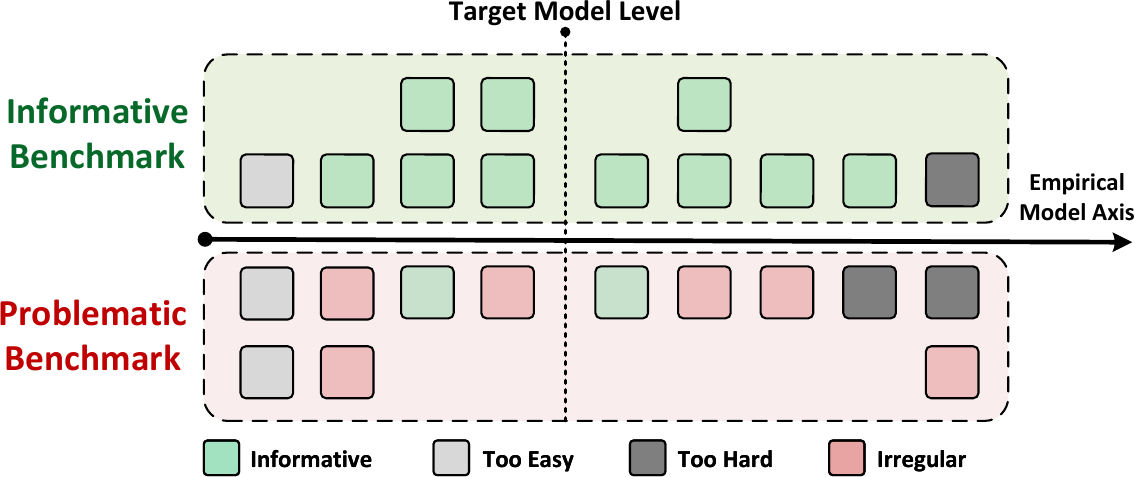}
  \caption{Schematic comparison of two hypothetical benchmarks with the same target-model score but different item compositions.}
  \label{fig:motivation}
\end{figure}

Large language model (LLM) benchmarks assess model capabilities through performance on individual benchmark items~\citep{Hendrycks21mmlu,Laskar24}. The usefulness of these assessments depends on whether the items serve their intended measurement purpose~\citep{AERA2014Standards,jo25}. Ambiguous questions, incorrect reference answers, or flawed scoring rules can introduce measurement error unrelated to the target capability~\citep{Gema2025,Truong2025}. Even correctly specified items may offer limited discrimination when their difficulty is poorly matched to the target model range~\citep{Vania21}. Aggregate scores and leaderboards summarize model performance but provide limited insight into these item-level properties~\citep{Rodriguez21,Jiang26}. Figure~\ref{fig:motivation} illustrates that two benchmarks with different item compositions can yield the same model score. Benchmark designers and evaluators therefore need a structural diagnosis framework that provides item-level evidence for assessing suitability and guiding further review, together with benchmark-level summaries for assessing reliability.

Existing methods for benchmark item analysis can be broadly grouped into content auditing and response-based analysis. \textit{Content auditing} identifies content and scoring defects~\citep{Gema2025,wang2026auto}, but content correctness alone does not establish an item's suitability for distinguishing among the target models. \textit{Response-based analysis} uses model responses to estimate item difficulty and discrimination or identify anomalies. Parametric estimates of item properties depend on assumptions about ability structure and response functions, as well as the ability range covered by calibration data~\citep{Vania21,Land2026}. Methods based on inter-item response associations can flag anomalous items~\citep{Truong2025,Hardy2026}, but their scores alone do not characterize how an item's discriminative behavior varies across the target model range. Some methods additionally require access to answer-option probabilities or manipulation of distractors, restricting their applicability across task formats and model interfaces~\citep{Song2026}. Accordingly, benchmark maintenance calls for interpretable item-level profiles that relate observed response behavior to an explicit empirical model reference and support both targeted review and benchmark-level analysis across task formats.

To this end, \textbf{B}enchmark \textbf{S}tructural \textbf{D}iagnosis via Ordered-Model \textbf{Probing} (\textbf{BSDProbe}) constructs interpretable item profiles from repeated responses along an empirically ordered model axis. Correctness and normalized entropy trajectories jointly characterize correctness transitions and their supporting response evidence through four dimensions: \textbf{boundary position}, \textbf{boundary width}, \textbf{boundary-signal validity}, and \textbf{order consistency}. These profiles identify response patterns that warrant further review and summarize benchmark-level transition coverage, providing evidence for assessing item suitability within the target model range. As a downstream application, profile-guided item selection achieves a mean relative improvement of 17.9\% in normalized discrimination ratio (NDR) over the strongest baseline per benchmark, with Qwen3 used for both selection and evaluation.

The main contributions are as follows:
\begin{itemize}
    \item BSDProbe provides a unified framework for benchmark structural diagnosis that converts repeated-response observations into interpretable four-dimensional item profiles through complementary correctness and normalized entropy trajectories. The profiles support automated item-level profiling and benchmark-level analysis conditional on an empirically ordered model axis.
    \item Experiments on six benchmarks spanning three task formats, with Qwen3 as the primary model axis, reveal distinct benchmark-level response structures. For downstream item selection, BSDProbe achieves higher mean NDRs than all five compared baselines on five of the six benchmarks.
    \item Repeated-response and profile-sensitivity analyses characterize the effects of response budgets and probe-model reduction on empirical observations and review decisions. Cross-axis comparisons reveal recurring benchmark-level patterns alongside variation in transition profiles and order-anomaly rates.
\end{itemize}

\begin{figure*}[ht]
  \centering
  \includegraphics[width=1.0\linewidth]{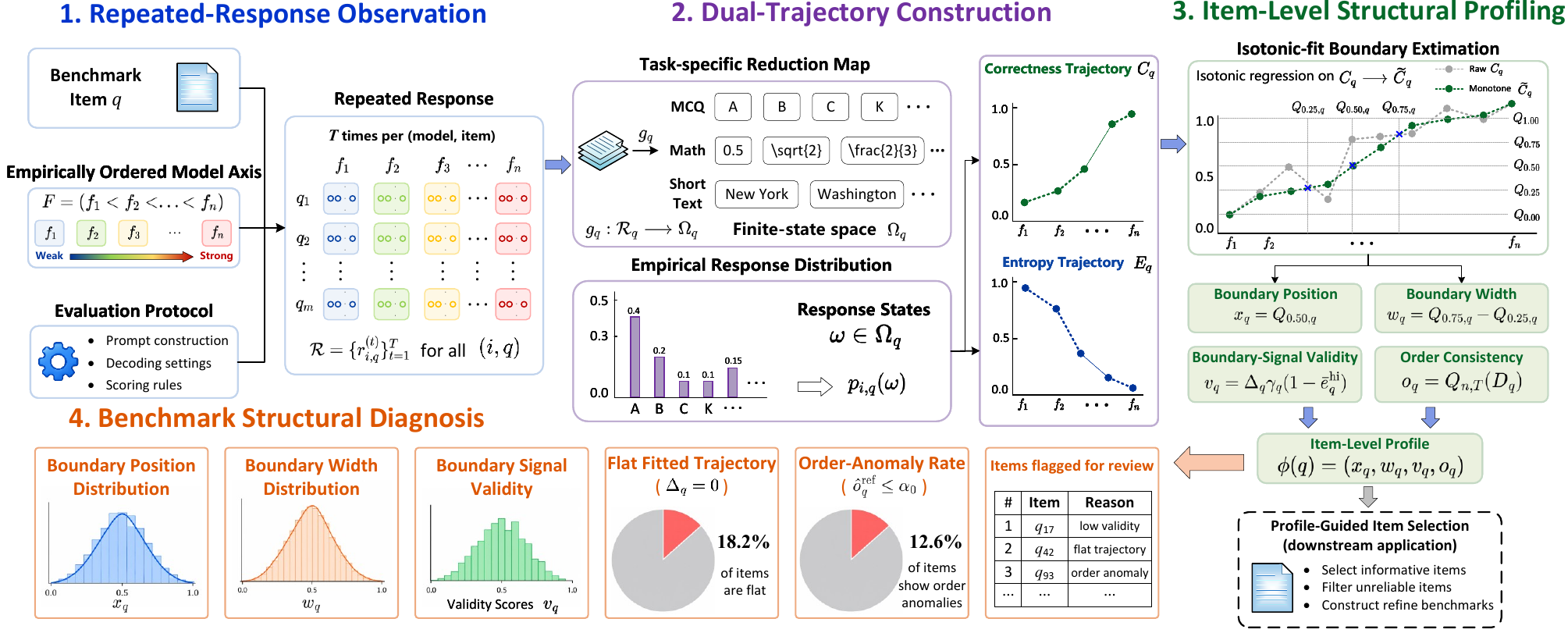}
  \caption{Overview of BSDProbe. Repeated responses along an empirically ordered model axis yield correctness and normalized entropy trajectories, which jointly support four-dimensional item profiles. Aggregating these profiles enables benchmark-level structural diagnosis, while order-consistency screening identifies items for further review.}
  \label{fig:framework}
\end{figure*}

\section{Related Work}
Benchmark auditing combines response-based screening with content verification to identify and investigate potentially flawed items. Fantastic Bugs~\citep{Truong2025} uses inter-item and item--total response associations to prioritize items for review. Signed isotonic scalability coefficients use inter-item isotonic regression to measure monotone associations across mixed response types~\citep{Hardy2026}. Concurrent probability-landscape auditing examines competition among answer options through option probabilities and distractor perturbations~\citep{Song2026}. These statistical signals complement expert reannotation and automated auditing, which examine items and scoring implementations for concrete defects~\citep{Gema2025,wang2026auto}. Benchmark Profiling uses gradient-based importance scoring and targeted parameter ablation to identify the abilities contributing to benchmark performance~\citep{kim25}. BSDProbe complements these audits with profiles of transition location and concentration along a specified model axis. Its order-review flags guide review without confirming content defects.

Item response theory (IRT) relates item informativeness to modeled ability for test-set comparison~\citep{Lalor16,Vania21,Rodriguez21}, while response-model fit can prioritize potentially flawed items for auditing~\citep{Land2026}. Nonparametric IRT allows more flexible response functions; monotone homogeneity models retain assumptions of unidimensionality, local independence, and monotonicity~\citep{sijtsma2002introduction}. Conditional on an empirical model axis, the isotonic fit to BSDProbe's correctness trajectory can be viewed as a discrete nonparametric item characteristic curve (ICC). Finite-state response distributions can differ in dispersion even when their correctness probabilities are identical. BSDProbe combines correctness and dispersion information for structural profiling. Repeated sampling improves estimation of response distributions at fixed model positions without expanding model coverage. BSDProbe conditions diagnosis on empirical axis ordering and coverage, without jointly estimating latent abilities.

Item-level response information also supports efficient evaluation. tinyBenchmarks and metabench use IRT-based estimation to reduce evaluation budgets~\citep{polo24,Kipnis25}, while Anchor Points selects representative items from model-confidence patterns~\citep{vivek24}. Amortized difficulty prediction supports adaptive evaluation~\citep{Truong25}, and MetaEval combines signal detection with item-response modeling to quantify discrimination and construct compact subsets~\citep{Wang26}. BSDProbe evaluates profile-guided item selection as a downstream application of structural diagnosis, examining both model discrimination and ranking agreement.

\section{Benchmark Structural Diagnosis via Ordered-Model Probing}
Assessing item suitability requires examining response variation across the target models. BSDProbe uses repeated responses along an empirically ordered model axis to support item-level profiling and benchmark structural diagnosis (Figure~\ref{fig:framework}).

\subsection{Observation Protocol and Trajectories}

For a benchmark item $q$, let $\mathcal{F}=(f_1,f_2,\dots,f_n)$ denote probe models ordered by empirical performance on the target task, with $f_1\prec f_2\prec\dots\prec f_n$. This \textbf{ordered model axis} provides a reference for interpreting cross-model response changes within the evaluated model range. For each model--item pair $(f_i,q)$, $T$ responses $\{r_{i,q}^{(t)}\}_{t=1}^{T}$ are collected under a fixed evaluation protocol specifying the prompt, decoding configuration, and scoring rules. These responses are modeled as independent and identically distributed conditional on the model, item, and evaluation protocol. Repeated sampling provides distributional information beyond a single realized response.

A task-specific reduction map $g_q:\mathcal{R}_q\rightarrow\Omega_q$ maps raw responses to answer states, where $\mathcal{R}_q$ is the raw response space and $\Omega_q$ is the task-specific state space. This representation accommodates multiple-choice answers, short-text answers, and mathematical reasoning outputs. The same reduction map and scoring rules are applied across model positions for each item, and the map must preserve correctness judgments. Let $\widehat{\Omega}_q$ denote the set of answer states observed across all probe models for item $q$, with $M_q=|\widehat{\Omega}_q|$. The subset $\widehat{\Omega}_q^\star\subseteq\widehat{\Omega}_q$ contains the states judged correct under the scoring rules. Applying $g_q$ to the repeated responses yields the empirical response distribution, defined for each $\omega\in\widehat{\Omega}_q$ as
\begin{equation}
    p_{i,q}(\omega)
    =\frac{1}{T}\sum_{t=1}^{T}
    \mathbf{1}\left[g_q\left(r_{i,q}^{(t)}\right)=\omega\right].
\end{equation}
Here, $p_{i,q}(\omega)$ is the empirical probability that model $f_i$ produces state $\omega$ on item $q$. States not observed at a model position have zero empirical probability.

Response dispersion is summarized by the \textbf{normalized entropy}:
\begin{equation}
e_{i,q}=\dfrac{-\sum_{\omega\in\widehat{\Omega}_q} p_{i,q}(\omega)\log p_{i,q}(\omega)}{\log M_q}
\end{equation}
with the convention $0\log 0=0$. This shared normalization provides a common reference across model positions within an item, with $e_{i,q}\in[0,1]$ and lower values indicating more concentrated responses. Response concentration alone does not indicate correctness. The \textbf{correctness mass} is therefore defined as
\begin{equation}
    c_{i,q}=\sum_{\omega\in\widehat{\Omega}_q^\star}p_{i,q}(\omega)
\end{equation}
where $c_{i,q}\in[0,1]$ denotes the empirical probability of a correct response. Along the ordered model axis, these measures form the normalized entropy trajectory $E_q=(e_{i,q})_{i=1}^{n}$ and the correctness trajectory $C_q=(c_{i,q})_{i=1}^{n}$. Together, $(E_q,C_q)$ provide the observational basis for item-level profiling.

\subsection{Item-Level Structural Profiles}

BSDProbe summarizes each item by a four-dimensional profile $\phi(q)=(x_q,w_q,v_q,o_q)$, comprising \textbf{boundary position}, \textbf{boundary width}, \textbf{boundary-signal validity}, and \textbf{order consistency}. Since $C_q$ may contain local reversals, an equal-weight isotonic fit $\widetilde C_q=(\tilde c_{i,q})_{i=1}^{n}$ provides a non-decreasing reference for correctness transitions~\citep{barlow1972statistical}, while the original trajectory is retained for order diagnosis. Fitting details are provided in Appendix~\ref{app:isotonic}.

For $n\ge2$, let $s_i=(i-1)/(n-1)$ denote normalized model positions and $\Delta_q=\tilde c_{n,q}-\tilde c_{1,q}$ the total fitted correctness gain. For $\Delta_q>0$ and $\alpha\in(0,1)$, define $Q_{\alpha,q}$ as the first normalized position where the linearly interpolated fit through $(s_i,\tilde c_{i,q})$ reaches $\tilde c_{1,q}+\alpha\Delta_q$. \textbf{Boundary position and width} are defined as
\begin{equation}
    x_q=Q_{0.50,q},
    \qquad
    w_q=Q_{0.75,q}-Q_{0.25,q}.
    \label{eq:boundary-position}
\end{equation}
Here, $x_q$ locates the midpoint of the fitted correctness gain, while $w_q$ measures the normalized model interval over which its central $50\%$ occurs. Smaller widths indicate more concentrated transitions.

When $\Delta_q=0$, the fitted trajectory does not identify a transition. An approximate position $x_q=h_{\mathcal B,\mathcal F}(a_q)$ is assigned using mean empirical correctness $a_q=\frac{1}{n}\sum_{i=1}^{n}c_{i,q}$, and $w_q=1$ indicates that no transition can be localized. The reference function $h_{\mathcal B,\mathcal F}$ uses items with positive fitted correctness gain from the same benchmark $\mathcal B$ and model axis $\mathcal F$ and returns positions in $[0,1]$ (Appendix~\ref{app:flat-position}). These assigned positions and widths are retained for complete item-level visualizations but excluded from summaries of observed correctness transitions.

\textbf{Boundary-signal validity} $v_q$ provides a response-based weight for interpreting the item's estimated boundary position $x_q$ and width $w_q$. Let $k_{i,q}$ denote the number of correct responses among $T$ observations at model position $i$ and $\theta_{i,q}$ the corresponding correctness probability for item $q$ under the evaluation protocol. Under a binomial sampling model with independent $\operatorname{Beta}(1,1)$ priors for the endpoint probabilities, posterior confidence in endpoint improvement is defined as
\begin{equation}
    \gamma_q=\Pr(\theta_{n,q}>\theta_{1,q}\mid k_{1,q},k_{n,q}).
\end{equation}
For $\Delta_q>0$, let $H_q=\{i:s_i\ge Q_{0.75,q}\}$ denote the model indices where at least $75\%$ of the item's fitted correctness gain has been attained. Their mean normalized entropy is $\bar e_q^{\mathrm{hi}}=\frac{1}{|H_q|}\sum_{i\in H_q}e_{i,q}$. The weight is defined as
\begin{equation}
    v_q=\Delta_q\gamma_q(1-\bar e_q^{\mathrm{hi}}).
    \label{eq:boundary-validity}
\end{equation}
The score satisfies $0\le v_q\le\Delta_q\le1$. Limited fitted correctness gain, weak posterior support for endpoint improvement, or dispersed high-side responses reduce this weight. When $\Delta_q=0$, $v_q=0$ without evaluating the high-side region. The endpoint posterior and computation of $\gamma_q$ are detailed in Appendix~\ref{app:boundary-validity}.

\textbf{Order consistency} $o_q$ assesses whether reversals in the original correctness trajectory are unusually pronounced under non-decreasing correctness probabilities along the empirical model axis. Under conditional binomial sampling, $k_{i,q}\sim\operatorname{Binomial}(T,\theta_{i,q})$ independently across model positions. The null hypothesis restricts these probabilities to $\mathcal M_n=\{\theta\in[0,1]^n:\theta_1\le\cdots\le\theta_n\}$. With a common response budget, the isotonic fit $\widetilde C_q$ is also the monotone binomial maximum-likelihood estimate. The resulting likelihood-ratio statistic quantifies the loss incurred by imposing monotonicity~\citep{tijmstra2013testing}:
\begin{equation}
    D_q=2T\sum_{i=1}^{n}\operatorname{KL}(\operatorname{Bern}(c_{i,q})\Vert\operatorname{Bern}(\tilde c_{i,q})).
\end{equation}
Here, $\operatorname{Bern}(p)$ denotes a Bernoulli distribution with success probability $p$, and $\operatorname{KL}$ denotes Kullback--Leibler divergence, using natural logarithms and zero-mass terms interpreted as zero. Larger values indicate a greater likelihood loss when the item's observed correct-response counts are fitted under the model-order constraint.

To assess whether this loss is unusually large under monotonicity, $D_q$ is calibrated against its null distribution. Because this distribution depends on unknown correctness probabilities, the theoretical order-consistency score takes the largest inclusive upper-tail probability over $\mathcal M_n$~\citep{dufour2006monte}:
\begin{equation}
    o_q=Q_{n,T}(D_q)=\sup_{\theta\in\mathcal M_n}\Pr_{\theta}\left(D_{n,T}^{\ast}\ge D_q\right).
\end{equation}
Here, $D_{n,T}^{\ast}$ is computed from independent binomial counts under $\theta$, using the same fitting and statistic rules. Small $o_q$ indicates evidence against monotonicity along the given axis. Under the fixed observation design, exact global calibration controls the per-item false-positive probability of the rule $o_q\le\alpha_o$ at the prespecified level $\alpha_o$.

In experiments, $o_q$ is approximated by a finite-reference Monte Carlo score $\widehat o_q^{\mathrm{ref}}$. Screening applies to all items, including those with flat isotonic fits, and is separate from boundary-signal validity. The derivation, calibration, and empirical validation are detailed in Appendix~\ref{app:profile_details}.

\subsection{Benchmark-Level Diagnosis}

At the benchmark level, distributions of $x_q$ and $w_q$ are summarized for items with $\Delta_q>0$, together with the distribution of $v_q$ across all items and the proportion of flat fitted trajectories. Weighting the position and width distributions by $v_q$ highlights transitions with stronger response support; normalized weighted distributions are reported only when the total weight is positive.

Items satisfying $\widehat o_q^{\mathrm{ref}}\le\alpha_o$ are flagged for further review. Their proportion is reported, and the flags are retained alongside item attributes and response trajectories to support targeted inspection. These summaries characterize benchmark response structure within the evaluated model range and provide evidence for benchmark maintenance.

\begin{figure*}[ht]
    \centering
    \includegraphics[width=0.95\linewidth]{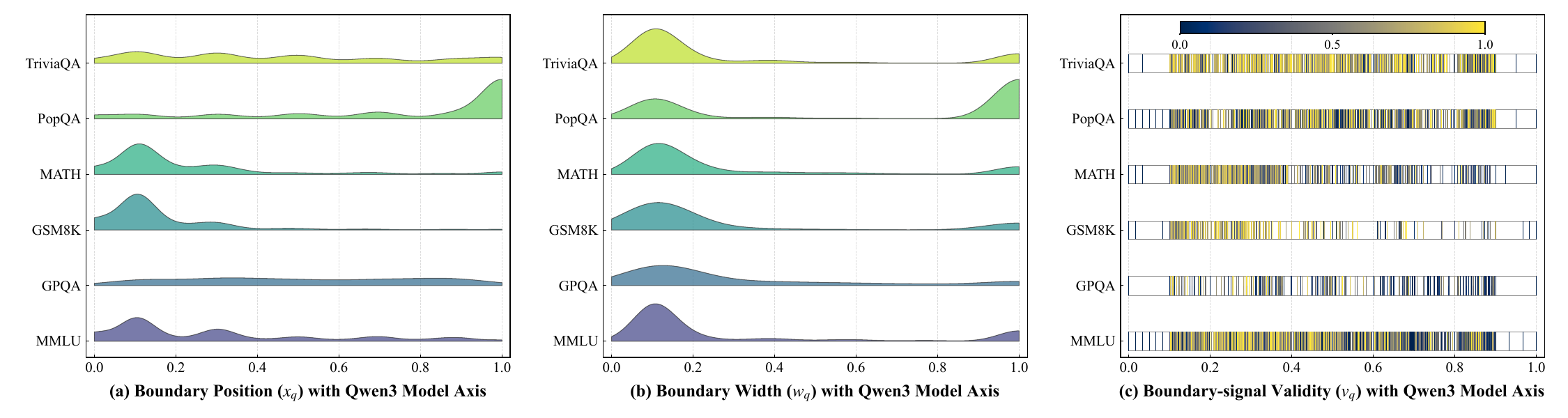}
    \caption{Benchmark structural profiles along the Qwen3 axis. Each row corresponds to one benchmark, showing \textbf{(a)} boundary-position distributions $x_q$, \textbf{(b)} boundary-width distributions $w_q$, and \textbf{(c)} boundary-signal validity $v_q$.}
    \label{fig:qwen3_bsd}
\end{figure*}

\begin{figure*}[ht]
    \centering
    \includegraphics[width=0.95\linewidth]{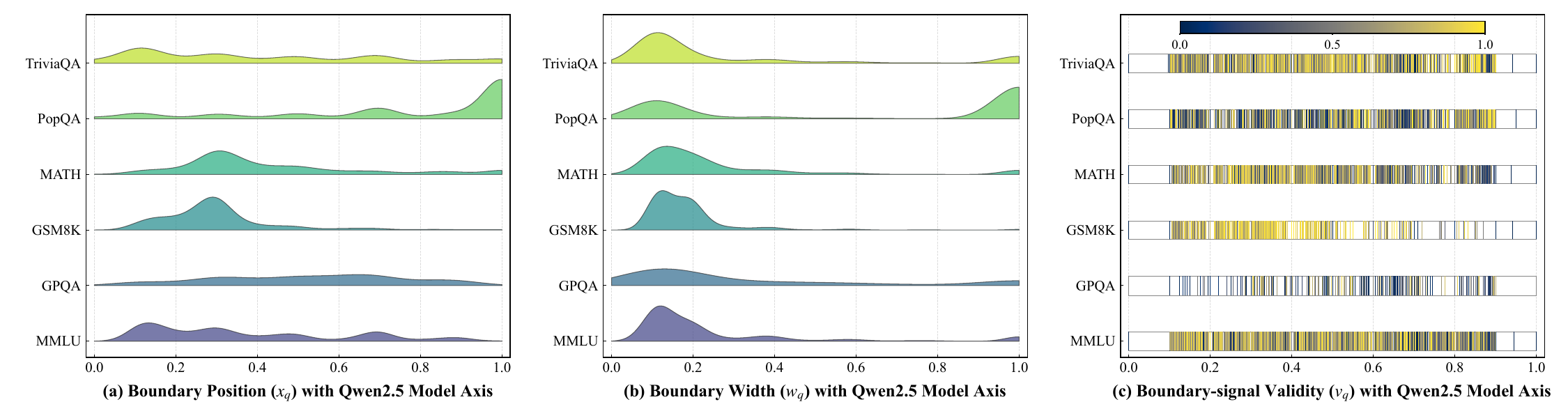}
    \caption{Benchmark structural profiles along the Qwen2.5 axis.}
    \label{fig:qwen25_bsd}
\end{figure*}

\begin{figure*}[ht]
    \centering
    \includegraphics[width=0.95\linewidth]{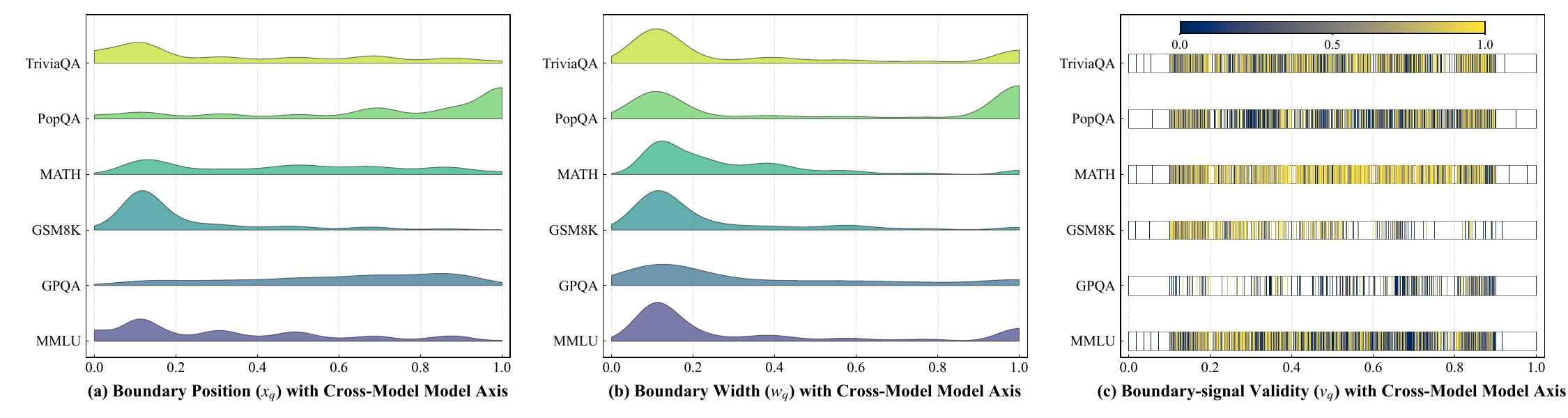}
    \caption{Benchmark structural profiles along the empirically ordered cross-model axis.}
    \label{fig:cross-model_bsd}
\end{figure*}

\section{Experiments}

\subsection{Experimental Setup}
To systematically evaluate BSDProbe, we define three research questions (RQs): 
\begin{itemize}
    \item \textbf{RQ1: Benchmark Structural Diagnosis.} What item-level and benchmark-level response structures does BSDProbe reveal?
    \item \textbf{RQ2: Observation Reliability and Axis Sensitivity.} How do sampling budgets and model-axis design affect the reliability and interpretation of structural diagnoses?
    \item \textbf{RQ3: Measurement Utility.} Does profile-guided item selection improve model discrimination relative to baseline selection methods?
\end{itemize}

Experiments use six public benchmarks spanning three task formats: multiple-choice questions (MCQ), represented by MMLU~\citep{Hendrycks21mmlu} and GPQA~\citep{Rein23gpqa}; mathematical reasoning (Math), represented by GSM8K~\citep{Cobbe21gsm8k} and MATH~\citep{Hendrycks21math}; and short-text question answering (QA), represented by PopQA~\citep{Mallen23popqa} and TriviaQA~\citep{Joshi17triviaqa}. Task-specific response mappings and scoring rules are applied consistently across model positions. Prompt templates, mapping rules, scoring procedures, and recorded decoding settings are provided in Appendix~\ref{app:reproducibility}.

Three model-axis settings are evaluated, each comprising six probe models. The Qwen3 series (0.6B--32B) serves as the primary within-family axis, while the Qwen2.5 series (0.5B--32B) and a heterogeneous axis comprising Llama, GLM, and DeepSeek-R1-Distill-Qwen models provide comparative settings. Models are ordered separately for each benchmark by empirical performance on its analyzed items, and the resulting order is shared across all items. Model rosters and ordering details are reported in Appendix~\ref{app:multi_axis_validation}. Diagnoses are interpreted conditional on the selected model axis and evaluation protocol.

Unless otherwise specified, analyses use $T=20$ responses per model--item pair. Boundary-signal validity $v_q$ is reported as a continuous weight, and items satisfying $\widehat o_q^{\mathrm{ref}}\le0.05$ are flagged for review using finite-reference Monte Carlo calibration (Appendix~\ref{app:profile_details}). For GPQA items whose missing historical option mappings prevent exact entropy normalization, validity weights are reported as bounds.

\subsection{Benchmark Structural Diagnosis}

This experiment applies BSDProbe along two within-family axes (Qwen3 and Qwen2.5) and a heterogeneous cross-model axis to characterize the benchmark-level structural profiles of six benchmarks. Cross-axis comparisons examine recurring response patterns and variation associated with the probe-model sets and their empirical orderings.

The structural profiles reveal distinct patterns of transition location and concentration across benchmarks (Figures~\ref{fig:qwen3_bsd}, \ref{fig:qwen25_bsd}, and~\ref{fig:cross-model_bsd}). Among positive-gain items on the Qwen3 axis, correctness transitions occur earlier on average for GSM8K and MATH than for GPQA and PopQA. Among items with positive fitted correctness gain, GPQA has the largest mean boundary width across all three axes, indicating less concentrated fitted correctness transitions. MATH has later mean boundary positions on the Qwen2.5 and cross-model axes than on Qwen3, illustrating variation across empirical model axes.

Boundary-signal validity complements these position and width patterns by describing the response evidence supporting the transitions. PopQA has the largest proportion of items with flat fitted correctness trajectories ($\Delta_q=0$) across all three axes, reaching 61.93\% on Qwen3. Among positive-gain items on this axis, PopQA and TriviaQA exhibit similarly concentrated transitions, but their mean validity weights are 0.439 and 0.699, respectively. Comparable transition concentration therefore does not imply comparable support from correctness gain, endpoint improvement evidence, and high-side response concentration.

\begin{table}[ht]
    \centering
    \small
    \caption{Order-anomaly rates (\%) across model axes. Each rate uses all analyzed items in the corresponding benchmark--axis condition.}
    \label{tab:order_anomaly_rates}
    \begin{tabular}{l|rrr}
        \toprule
        Benchmark & Qwen3 & Qwen2.5 & Cross-model \\
        \midrule
        MMLU     & 31.80 & 26.78 & 31.04 \\
        GPQA     & 56.25 & 48.21 & 57.59 \\
        GSM8K    &  8.87 &  3.79 & 28.13 \\
        MATH     &  8.18 &  4.25 &  7.27 \\
        PopQA    & 20.04 & 20.34 & 27.48 \\
        TriviaQA & 25.28 & 25.82 & 36.96 \\
        \bottomrule
    \end{tabular}
\end{table}

Table~\ref{tab:order_anomaly_rates} reports the proportions of items flagged for order-focused review. GPQA has the highest rate on all three axes, ranging from 48.21\% to 57.59\%, making order-focused review a recurring priority across the tested axis settings. MATH has comparatively low rates of 4.25--8.18\%, whereas GSM8K varies substantially across axes: its order-anomaly rate is 3.79\% on Qwen2.5 and 28.13\% on the cross-model axis. Recurring benchmark-level patterns coexist with axis-dependent variation.

Inspection of items selected through order screening reveals concrete content and scoring issues. Appendix~\ref{app:representative_cases} presents one illustrative case from each of the six benchmarks, together with the question, scoring reference, model responses, and diagnostic evidence. Subsequent review identifies equivalent or duplicate options, incorrect or mismatched reference answers, an underspecified entity reference, and a reference-extraction error in the MATH preprocessing pipeline used in the experiments. These cases illustrate how BSDProbe can prioritize items for targeted review and help identify issues in benchmark content and evaluation procedures.

\subsection{Observation Reliability and Axis Sensitivity}

\textbf{Repeated-response reliability} is assessed using 50 stored Qwen3-8B responses per item across the six benchmarks. Mean within-item Jensen--Shannon divergence (JSD) between disjoint response blocks, computed in each item's answer-state space, decreases as block size increases from 5 to 25. Cross-item comparisons use a shared space of correct, valid-but-incorrect, and invalid outcomes for both within-item and cross-item JSD. At 20 responses per block, mean cross-item JSD exceeds mean within-item JSD by 0.0648--0.2853 across benchmarks, with all 95\% bootstrap confidence intervals for these differences above zero (Appendix~\ref{app:observation_reliability}). These results support repeatable, item-dependent response behavior under the tested protocol.

\textbf{Profile sensitivity} is assessed relative to the six-model setting with $T=20$, using order-score calibration matched to each model count and response budget (Appendix~\ref{app:axis_sensitivity}). On Qwen3, subsampling $T=15$ responses yields mean order-review agreement of 89.8--98.3\% with the reference setting over 100 repetitions. Among items with positive fitted gain in both conditions, mean absolute changes in boundary position and width are at most 0.029 and 0.041, respectively. At $T=20$, endpoint-preserving three-model subaxes yield agreement of 65.6--94.8\% across the tested subaxes and benchmarks. Retaining both endpoints therefore does not by itself preserve item-level review decisions.

\subsection{Measurement Utility}
This experiment evaluates whether item-level structural profiles support subset selection that improves model discrimination while preserving model rankings. BSDProbe combines boundary-position stratification with sampling weights $(1-w_q)v_q$ after order screening. Comparisons include \textit{PSN-IRT}~\citep{zhou26}, \textit{TinyBenchmarks}~\citep{polo24}, \textit{Anchor Points}~\citep{vivek24}, \textit{Bayesian}~\citep{xiao25}, and uniform \textit{Random} sampling, using the same candidate pool and matched subset sizes within each benchmark.

For a reference axis $a$ with $n_a$ models, let $\mathrm{Score}^{(a)}_i(\mathcal A)$ denote the principal-answer accuracy of its $i$th model on an item set $\mathcal A$, and define the adjacent-model gap as $\Delta_i^{(a)}(\mathcal A)=\mathrm{Score}^{(a)}_{i+1}(\mathcal A)-\mathrm{Score}^{(a)}_i(\mathcal A)$. The \textbf{normalized discrimination ratio (NDR)} of a selected subset $\mathcal S$ is
\begin{equation}
    \mathrm{NDR}^{(a)}(\mathcal S)=\frac{\sum_{i=1}^{n_a-1}[\Delta_i^{(a)}(\mathcal S)]_+}{\sum_{i=1}^{n_a-1}|\Delta_i^{(a)}(\mathcal Q^{(a)})|},
\end{equation}
where $[z]_+=\max(0,z)$ and $\mathcal Q^{(a)}$ is the full scoreable item set for the reference axis. With a positive denominator, NDR above one indicates greater positive adjacent-model score separation than the full benchmark. Kendall $\tau_b$ additionally measures agreement between model rankings on the subset and full benchmark.

\begin{table*}[ht]
\centering
\caption{Subset discriminability measured by NDR for 100-item subsets, with all NDR values evaluated on the Qwen3 reference axis ($\uparrow$ indicates higher discrimination). BSDProbe labels indicate the axis used for subset selection. Values are reported as mean $\pm$ standard deviation over repeated subset sampling. Bold marks the best result in each row.}
\label{tab:measurement_ndr_qwen3}
\resizebox{\textwidth}{!}{
\setlength{\tabcolsep}{2pt}
\begin{tabular}{l|ccccc|ccc}
\toprule
\textbf{Benchmark} & Random & \makecell{Tiny\\Benchmarks} & \makecell{Anchor\\Points} & Bayesian & PSN-IRT & \makecell{BSDProbe\\(Qwen3)} & \makecell{BSDProbe\\(Qwen2.5)} & \makecell{BSDProbe\\(Cross-model)} \\
\midrule
MMLU     & $1.00_{\pm 0.09}$ & $1.48_{\pm 0.06}$ & $0.69_{\pm 0.06}$ & $1.03_{\pm 0.15}$ & $1.65_{\pm 0.12}$ & $\mathbf{2.04_{\pm 0.06}}$ & $1.27_{\pm 0.10}$ & $1.35_{\pm 0.09}$ \\
GPQA     & $0.99_{\pm 0.19}$ & $2.23_{\pm 0.14}$ & $0.80_{\pm 0.08}$ & $1.26_{\pm 0.21}$ & $1.08_{\pm 0.00}$ & $\mathbf{2.90_{\pm 0.14}}$ & $1.75_{\pm 0.12}$ & $1.53_{\pm 0.15}$ \\
\midrule
GSM8K    & $1.09_{\pm 0.24}$ & $0.57_{\pm 0.12}$ & $1.44_{\pm 0.12}$ & $1.77_{\pm 0.00}$ & $2.45_{\pm 0.18}$ & $\mathbf{2.55_{\pm 0.18}}$ & $1.93_{\pm 0.18}$ & $1.53_{\pm 0.15}$ \\
MATH     & $1.12_{\pm 0.14}$ & $1.08_{\pm 0.07}$ & $1.51_{\pm 0.11}$ & $1.83_{\pm 0.13}$ & $0.44_{\pm 0.00}$ & $\mathbf{2.07_{\pm 0.11}}$ & $1.39_{\pm 0.10}$ & $1.23_{\pm 0.14}$ \\
\midrule
PopQA    & $1.00_{\pm 0.36}$ & $0.36_{\pm 0.08}$ & $1.23_{\pm 0.09}$ & $3.10_{\pm 0.46}$ & $4.42_{\pm 0.00}$ & $\mathbf{6.06_{\pm 0.19}}$ & $4.20_{\pm 0.26}$ & $3.28_{\pm 0.28}$ \\
TriviaQA & $0.96_{\pm 0.09}$ & $\mathbf{1.68_{\pm 0.03}}$ & $0.76_{\pm 0.04}$ & $1.28_{\pm 0.09}$ & $1.40_{\pm 0.06}$ & $\mathbf{1.68_{\pm 0.03}}$ & $1.30_{\pm 0.06}$ & $1.04_{\pm 0.05}$ \\
\bottomrule
\end{tabular}
}
\end{table*}

\begin{table*}[ht]
\centering
\caption{Ranking agreement measured by Kendall $\tau_b$ for 100-item subsets evaluated on the Qwen3 reference axis ($\uparrow$ indicates higher agreement). BSDProbe labels indicate the axis used for subset selection. Values are reported as mean $\pm$ standard deviation over repeated subset sampling. Bold marks the best result in each row.}
\label{tab:measurement_kendall_tau_b_qwen3}
\resizebox{\textwidth}{!}{
\setlength{\tabcolsep}{2pt}
\begin{tabular}{l|ccccc|ccc}
\toprule
\textbf{Benchmark} & Random & \makecell{Tiny\\Benchmarks} & \makecell{Anchor\\Points} & Bayesian & PSN-IRT & \makecell{BSDProbe\\(Qwen3)} & \makecell{BSDProbe\\(Qwen2.5)} & \makecell{BSDProbe\\(Cross-model)} \\
\midrule
MMLU     & $0.943_{\pm 0.092}$ & $0.874_{\pm 0.068}$ & $0.997_{\pm 0.011}$ & $0.906_{\pm 0.088}$ & $0.983_{\pm 0.042}$ & $\mathbf{1.000_{\pm 0.000}}$ & $0.990_{\pm 0.016}$ & $\mathbf{1.000_{\pm 0.000}}$ \\
GPQA     & $0.895_{\pm 0.148}$ & $\mathbf{1.000_{\pm 0.000}}$ & $0.820_{\pm 0.127}$ & $0.599_{\pm 0.154}$ & $\mathbf{1.000_{\pm 0.000}}$ & $\mathbf{1.000_{\pm 0.000}}$ & $0.973_{\pm 0.056}$ & $0.976_{\pm 0.042}$ \\
\midrule
GSM8K    & $0.845_{\pm 0.166}$ & $0.741_{\pm 0.099}$ & $0.952_{\pm 0.047}$ & $\mathbf{0.966_{\pm 0.000}}$ & $0.941_{\pm 0.043}$ & $\mathbf{0.966_{\pm 0.000}}$ & $0.843_{\pm 0.087}$ & $0.948_{\pm 0.047}$ \\
MATH     & $0.783_{\pm 0.106}$ & $0.860_{\pm 0.074}$ & $0.910_{\pm 0.105}$ & $0.873_{\pm 0.090}$ & $0.775_{\pm 0.000}$ & $\mathbf{1.000_{\pm 0.000}}$ & $0.753_{\pm 0.167}$ & $0.720_{\pm 0.237}$ \\
\midrule
PopQA    & $0.893_{\pm 0.076}$ & $0.649_{\pm 0.177}$ & $0.872_{\pm 0.157}$ & $0.896_{\pm 0.106}$ & $\mathbf{1.000_{\pm 0.000}}$ & $\mathbf{1.000_{\pm 0.000}}$ & $0.967_{\pm 0.054}$ & $0.983_{\pm 0.054}$ \\
TriviaQA & $0.876_{\pm 0.086}$ & $0.909_{\pm 0.055}$ & $0.987_{\pm 0.042}$ & $0.930_{\pm 0.067}$ & $0.997_{\pm 0.011}$ & $\mathbf{1.000_{\pm 0.000}}$ & $\mathbf{1.000_{\pm 0.000}}$ & $\mathbf{1.000_{\pm 0.000}}$ \\
\bottomrule
\end{tabular}
}
\end{table*}

The main comparison uses Qwen3 as the reference axis and evaluates BSDProbe subsets selected using Qwen3, Qwen2.5, and cross-model. Results for the Qwen2.5 and cross-model reference axes are reported in Appendix~\ref{app:cross_axis_transfer}.

On the Qwen3 reference axis, BSDProbe selected using Qwen3 achieves the highest mean NDR on five of the six benchmarks (Table~\ref{tab:measurement_ndr_qwen3}). Relative to the strongest baseline on each benchmark, its mean relative NDR improvement across the six benchmarks is 17.9\%, with gains of 24.1\% on MMLU, 30.1\% on GPQA, and 37.1\% on PopQA. On TriviaQA, BSDProbe achieves an NDR of 1.68, matching the best baseline, TinyBenchmarks, at the reported precision. Overall, these results indicate that profile-guided selection can concentrate positive adjacent-model score separation into compact subsets.

The discrimination gains are accompanied by strong ranking preservation (Table~\ref{tab:measurement_kendall_tau_b_qwen3}). Under Qwen3 selection, BSDProbe achieves the highest or jointly highest mean Kendall $\tau_b$ on every benchmark, reaching 1.000 on five benchmarks and 0.966 on GSM8K. On GPQA and PopQA, it combines perfect ranking agreement in the tested runs with NDRs of 2.90 and 6.06, respectively. The joint results support the downstream value of structural profiles: the selected subsets amplify positive score separation while retaining close agreement with the full-benchmark model rankings.

Cross-axis selection also retains useful discrimination signals. Subsets selected using Qwen2.5 or cross-model achieve mean NDRs above one on all six benchmarks when evaluated on Qwen3. Ranking agreement remains particularly strong on MMLU, GPQA, and TriviaQA, with mean Kendall $\tau_b$ values of at least 0.990, 0.973, and 1.000, respectively. Transfer is nevertheless benchmark-dependent: on MATH, $\tau_b$ decreases to 0.753 and 0.720 under Qwen2.5 and cross-model selection, while NDR remains above one. Structural profiles therefore support discrimination beyond the selection models, although preserving the reference-model ranking depends on the benchmark and selection axis.

\section{Conclusion}
BSDProbe provides an item-level framework for diagnosing benchmark response structures and supporting benchmark maintenance. Repeated responses along an empirically ordered model axis yield correctness and normalized entropy trajectories, which support four-dimensional profiles of correctness transitions, response evidence, and ordering departures. Experiments on six benchmarks reveal response structures and sensitivity to model-axis design. With Qwen3 used for selection and evaluation, profile-guided selection achieves a 17.9\% mean relative NDR improvement over the strongest baseline per benchmark, with perfect ranking agreement on five benchmarks. The profiles can guide item review, benchmark revision, and compact subset construction. Future directions include adaptive probing and integration with content verification for benchmark maintenance.

\section*{Limitations}
BSDProbe's diagnoses depend on the coverage and ordering of the probe models and the evaluation protocol. Changes in these conditions can alter estimated transitions and item-level review priorities. Repeated sampling improves estimation at existing model positions, while correctness transitions outside the covered range remain unobserved. Task-specific answer normalization and scoring rules also influence the response distributions and dispersion estimates. Structural profiles consequently describe item behavior under these operational choices. Anomalies can arise from item defects, model specialization, or scoring artifacts, and identifying their causes requires examining the items, responses, and scoring decisions together. Order screening uses finite-reference Monte Carlo calibration to approximate the monotonicity null. Uniform false-positive control over the continuous null is not established, and the calibration treats the model ordering estimated from the analyzed items as fixed. Reported order-anomaly rates should therefore be interpreted within the stated observation design and calibration. The downstream experiments demonstrate selection utility across the tested benchmarks and model axes, with discrimination gains and ranking agreement varying according to the selection and reference models.

\bibliography{BSDProbe}

%%%%%%%%%%%%%%%%%%%%%%%%%%%%%%%%%%%%%%%%%%%%%%%%%%%%%%%%%%%%

\newpage
\appendix

\section{Isotonic Regression of Correctness Trajectories}
\label{app:isotonic}

For each item $q$, the empirical correctness trajectory $C_q=(c_{i,q})_{i=1}^{n}$ is fitted by a non-decreasing sequence. With a common response budget across model positions, BSDProbe applies equal-weight isotonic regression:
\begin{equation}
    \widetilde C_q
    =\underset{0\le z_1\le\cdots\le z_n\le1}
    {\operatorname{arg\,min}}
    \sum_{i=1}^{n}(c_{i,q}-z_i)^2.
\end{equation}
The strictly convex objective and convex constraint set yield a unique fitted trajectory $\widetilde C_q=(\tilde c_{i,q})_{i=1}^{n}$. The fit provides a non-decreasing reference for boundary characterization along the empirical model axis.

The fit is computed separately for each item using the pool-adjacent-violators algorithm (PAVA). Each block represents a contiguous segment of the ordered model axis whose positions share a fitted correctness level for that item. Initially, each model position forms a singleton block with empirical correctness $c_{i,q}$. When neighboring blocks have decreasing correctness levels along the model ordering, they are pooled into a merged block $\mathcal G$ with fitted level $\bar c_{\mathcal G,q}=\frac{1}{|\mathcal G|}\sum_{i\in\mathcal G}c_{i,q}$. Since each model contributes $T$ responses, this mean equals the proportion of correct responses pooled across the model positions in $\mathcal G$. Pooling continues until the block levels are non-decreasing, and each position $i\in \mathcal G$ receives the fitted value $\tilde c_{i,q}=\bar c_{\mathcal G,q}$.

A stack-based implementation requires $O(n)$ time and $O(n)$ storage per item. Equal fitted values form plateaus along the model axis. Pooling may produce a constant trajectory, indicating zero fitted correctness gain within the evaluated model range, even when the original trajectory contains substantial reversals. Both $C_q$ and $\widetilde C_q$ are retained for order diagnosis. The fitted trajectory determines the correctness gain and, when this gain is positive, the boundary position and width and high-side region used in subsequent profiling.

\section{Approximate Positioning without Fitted Correctness Gain}
\label{app:flat-position}

When $\Delta_q=0$, the fitted correctness trajectory does not identify a transition position. BSDProbe assigns an approximate position using the item's mean empirical correctness and reference items with positive fitted correctness gain from the same benchmark $\mathcal B$ and model axis $\mathcal F$.

For each item $q'$, let $a_{q'}=\frac{1}{n}\sum_{i=1}^{n}c_{i,q'}$ denote its mean empirical correctness. Let $u_1<\cdots<u_m$ be the distinct mean-correctness values among reference items satisfying $\Delta_{q'}>0$. At each value $u_j$, the reference position is
\begin{equation}
    b_j=
    \operatorname{median}
    \left\{
        Q_{0.50,q'}:
        q'\in\mathcal B,\Delta_{q'}>0,\ a_{q'}=u_j
    \right\}.
\end{equation}
The median combines transition positions from reference items with the same mean correctness. With a common model count and response budget, these items can be grouped by their total correct-response counts, avoiding floating-point comparisons of empirical means.

Two endpoint reference points specify the positioning conventions for empirically all-incorrect and all-correct items: $(u_0,b_0)=(0,1)$ and $(u_{m+1},b_{m+1})=(1,0)$. For $a\in[u_j,u_{j+1}]$, the reference function is defined by linear interpolation:
\begin{equation}
    h_{\mathcal B,\mathcal F}(a)
    =b_j+
    \frac{a-u_j}{u_{j+1}-u_j}
    (b_{j+1}-b_j).
\end{equation}
Each interpolated position is a convex combination of reference positions in $[0,1]$, ensuring $h_{\mathcal B,\mathcal F}(a)\in[0,1]$. If no reference items have positive fitted correctness gain, only the endpoint reference points remain, giving $h_{\mathcal B,\mathcal F}(a)=1-a$.

For an item with $\Delta_q=0$, the assigned position is $x_q=h_{\mathcal B,\mathcal F}(a_q)$, with $w_q=1$ and $v_q=0$. Similar mean correctness does not uniquely determine a transition position, so this assignment provides an empirical positioning convention rather than an observed transition. Assigned positions and widths are retained for complete item-level visualizations but excluded from summaries of observed correctness transitions. Order consistency remains evaluated from the item's original correct-response counts.

\section{Posterior Confidence for Boundary-Signal Validity}
\label{app:boundary-validity}

To assess correctness improvement along the empirical model axis, posterior confidence in the gain between its endpoint models is computed. For each item $q$, let $k_{1,q}$ and $k_{n,q}$ denote the number of correct responses among $T$ repeated observations for the first and last models, respectively, and let $\theta_{1,q}$ and $\theta_{n,q}$ denote their corresponding correctness probabilities under the evaluation protocol. Assuming independent binomial observations and uniform $\operatorname{Beta}(1,1)$ priors, the posterior distributions are
\begin{equation}
    \begin{aligned}
        \theta_{1,q}\mid k_{1,q}
        &\sim\operatorname{Beta}
        (k_{1,q}+1,T-k_{1,q}+1),\\
        \theta_{n,q}\mid k_{n,q}
        &\sim\operatorname{Beta}
        (k_{n,q}+1,T-k_{n,q}+1).
    \end{aligned}
\end{equation}

The posterior confidence that correctness increases between the endpoints is
\begin{equation}
    \gamma_q=
    \Pr(\theta_{n,q}>\theta_{1,q}
    \mid k_{1,q},k_{n,q}).
\end{equation}
This probability can be computed by numerical integration of the two Beta posterior distributions or approximated by Monte Carlo sampling. In the latter case, independent draws are taken from each posterior, and $\gamma_q$ is estimated by the fraction of paired draws satisfying $\theta_{n,q}>\theta_{1,q}$. Equal endpoint counts yield identical posteriors and $\gamma_q=0.5$, representing neutral directional evidence. The same calculation applies to different repetition budgets $T$, with sampling uncertainty reflected in the posterior distributions.

BSDProbe combines $\gamma_q$, computed from the original endpoint counts, with the fitted gain $\Delta_q$ and high-side response concentration to obtain $v_q$. The resulting response-based weight supports interpretation of the estimated boundary position and width. Ordering departures across the full trajectory are assessed separately by $o_q$.

\section{Likelihood-Ratio Calibration for Order Consistency}
\label{app:profile_details}

\subsection{Likelihood-Ratio Statistic}

Conditional on a fixed empirical model axis and evaluation protocol, the correct-response counts for item $q$ follow independent binomial distributions with a common response budget $T$. For a candidate correctness vector $z=(z_1,\ldots,z_n)\in[0,1]^n$, the log likelihood is
\begin{equation}
    \ell_q(z)
    =\sum_{i=1}^{n}
    \left[
        k_{i,q}\log z_i
        +(T-k_{i,q})\log(1-z_i)
    \right],
\end{equation}
where binomial coefficients are omitted because they cancel in the likelihood ratio. Natural logarithms are used, and zero-count terms are interpreted as zero.

The unrestricted maximum-likelihood estimate is $C_q$, with $c_{i,q}=k_{i,q}/T$. Under monotonicity, each pooled block $\mathcal G$ has fitted probability $\tilde c_{i,q}=\sum_{j\in \mathcal G}k_{j,q}/(T|\mathcal G|)$ for $i\in \mathcal G$. With a common response budget, this equals the arithmetic mean of the block's empirical correctness values. The monotone binomial maximum-likelihood estimate therefore coincides with the equal-weight isotonic fit $\widetilde C_q$~\citep{barlow1972statistical}; its computation is described in Appendix~\ref{app:isotonic}.

The likelihood-ratio statistic compares the unrestricted and monotone fits~\citep{tijmstra2013testing}:
\begin{equation}
    \begin{aligned}
        D_q
        &=2\left[\ell_q(C_q)-\ell_q(\widetilde C_q)\right]\\
        &=2T\sum_{i=1}^{n}
        \operatorname{KL}\!\left(
            \operatorname{Bern}(c_{i,q})
            \Vert
            \operatorname{Bern}(\tilde c_{i,q})
        \right).
    \end{aligned}
\end{equation}
For Bernoulli probabilities $u$ and $v$, the divergence is $u\log(u/v)+(1-u)\log((1-u)/(1-v))$, using continuous endpoint conventions. The statistic is non-negative and equals zero for a non-decreasing empirical trajectory. A flat isotonic fit can still yield $D_q>0$ when the original item responses contain reversals.

\subsection{Calibration and Screening}

The monotonicity null has parameter space $\mathcal M_n=\{\theta\in[0,1]^n:\theta_1\le\cdots\le\theta_n\}$. For $\theta\in\mathcal M_n$, let $D_{n,T}^{\ast}$ denote the statistic obtained from independent counts $k_i^\ast\sim\operatorname{Binomial}(T,\theta_i)$, with isotonic fitting repeated for each generated trajectory. Because its distribution depends on $\theta$, the theoretical calibration takes the largest inclusive upper-tail probability over the null~\citep{dufour2006monte}:
\begin{equation}
    Q_{n,T}(d)
    =\sup_{\theta\in\mathcal M_n}
    \Pr_{\theta}\left(D_{n,T}^{\ast}\ge d\right).
\end{equation}
The theoretical score is $o_q=Q_{n,T}(D_q)$. The inclusive comparison accounts for ties arising from discrete counts; in particular, $D_q=0$ gives $o_q=1$.

Under exact global calibration, the rule $o_q\le\alpha_o$ controls the per-item false-positive probability. Let $d_{\alpha_o}$ be the smallest attainable statistic satisfying $Q_{n,T}(d_{\alpha_o})\le\alpha_o$. Since $Q_{n,T}$ is non-increasing, every $\theta_0\in\mathcal M_n$ satisfies
\begin{equation}
    \Pr_{\theta_0}(o_q\le\alpha_o)=\Pr_{\theta_0}(D_q\ge d_{\alpha_o})\le Q_{n,T}(d_{\alpha_o})\le\alpha_o.
\end{equation}
If no attainable statistic satisfies the condition, the rejection set is empty. This guarantee applies to individual items under the stated sampling model and a prespecified model ordering. It does not provide benchmark-wide false-discovery control.

\subsection{Shared Calibration and Numerical Approximation}

Items with the same $n$, $T$, and fitting and statistic rules can share $Q_{n,T}$ because maximization over $\mathcal M_n$ accounts for their unknown correctness probabilities. Experiments approximate this calibration using a prespecified finite family $\Theta_{\mathrm{ref}}\subset\mathcal M_n$. For the primary setting with $n=6$ and $T=20$, 217 monotone reference vectors are used, including flat, linearly increasing, and piecewise-constant profiles.

For each reference vector $\theta$, $B=200000$ independent binomial count vectors are generated and refitted to obtain statistics $D_{\theta}^{\ast(b)}$ using the same procedure as observed items. Let $\rho(d)$ denote rounding to ten decimal places for consistent tail comparisons. The implemented score is
\begin{equation}
    \widehat o_q^{\mathrm{ref}}
    =
    \max_{\theta\in\Theta_{\mathrm{ref}}}
    \frac{
        1+\sum_{b=1}^{B}
        \mathbf{1}\left[
            \rho(D_{\theta}^{\ast(b)})\ge\rho(D_q)
        \right]
    }{B+1}.
\end{equation}

The inclusive comparison retains ties, and the add-one correction gives a minimum score of $1/(B+1)$. The frozen calibration table supports lookup without further simulation for individual items; its reference vectors and simulation settings are recorded for reproducibility.

Items with $\widehat o_q^{\mathrm{ref}}\le0.05$ are flagged for review. Finite-reference coverage and Monte Carlo uncertainty prevent this approximation from inheriting the exact score's full-null guarantee. Its false-flag behavior is assessed empirically using independent simulations of monotone correctness profiles.

\section{Experimental Reproducibility}
\label{app:reproducibility}

All experiments were conducted under fixed model versions, prompt templates, decoding configurations, random seeds, and task-specific state reduction rules to ensure reproducibility. Model inference was performed on a local vLLM framework (v0.8.5.post1), primarily on a single NVIDIA A800 GPU. The full experiments required approximately 200 to 250 GPU-hours. Random seeds were consistently set to 42. For item-level measurements, multiple repeated responses were collected to construct empirical distributions, while for comparisons across models or axes, a fixed item budget and repeated runs were used. Precomputed item-level statistics were reused for parameter sweeps and metric recalculation.

\textbf{Default configuration:} Temperature was set to 0.7, top-p to 0.8, presence penalty to 2, and top-k to 20. Maximum tokens followed the model-serving default unless otherwise required by task-specific answer extraction. For Qwen3 models, ``thinking'' mode was disabled.

\textbf{Task-specific state reduction and evaluation:} Model outputs were mapped to finite task-specific state spaces via a deterministic reduction function \(g_q\), differing by task type:

\begin{itemize}
    \item \textbf{Multiple-Choice (MMLU, GPQA):} Responses were reduced to discrete options (A/B/C/D) using JSON extraction, regex matching, and fallback to the last capital letter. Repeated outputs mapped to the same answer state are counted together. Evaluation compared normalized answers to reference answers in a case-insensitive manner.
    \item \textbf{Mathematical reasoning (GSM8K, MATH):} Responses were parsed for numeric and LaTeX answers. Fractions were converted to decimals, units removed, and multiple equivalence layers applied: string match, numeric equivalence (including percentage scaling), and symbolic equivalence via SymPy. Correctness is assessed using string, numeric, and symbolic equivalence to the reference answer.
    \item \textbf{Short-text Question Answering (PopQA, TriviaQA):} Responses were reduced by extracting the last line or content after ``ANSWER:''. Text was normalized (lowercase, punctuation removed, articles stripped, symbols unified). Invalid responses (e.g., ``I don't know'') are retained as a separate state and scored as incorrect. Reference-answer aliases are used in correctness scoring. Evaluation was based on normalized exact or substring matching.
\end{itemize}

\textbf{State distribution construction:} For all task types, the empirical response distribution $p_{i,q}$ was obtained by collecting repeated responses, applying the task-specific reducer $g_q$, counting the mapped answer states, and computing probabilities. Correctness mass $c_{i,q}$ is computed as the sum of probabilities of states judged correct under the benchmark scoring rules.

\section{Empirical Ordering of Model Axes}
\label{app:multi_axis_validation}

BSDProbe uses an empirically ordered model axis as the reference for item-level and benchmark-level diagnosis. The Qwen3 and Qwen2.5 settings use within-family candidate sets, while the heterogeneous setting combines models from multiple families. Each setting retains the same candidate models across benchmarks, with ordering determined separately for each benchmark.

Models are ranked by principal-answer accuracy on the shared set of analyzed items with complete, scoreable responses across all candidate models. This accuracy is the fraction of items for which a model's stored principal answer is judged correct under the benchmark's scoring rules. Ties retain the candidate-roster order, and the resulting ordering is applied to every item in that benchmark. The ordering is estimated from the same observations used for item profiling. It therefore provides an empirical reference within the evaluated model range, with diagnoses interpreted relative to the resulting benchmark-specific axis.

\textbf{Ordered Model Axes:} We evaluate three axes: (1) the primary Qwen3 series (0.6B to 32B); (2) the comparative Qwen2.5 series (0.5B to 32B); (3) a heterogeneous cross-model setting including LLaMA, GLM, Qwen, and DeepSeek models.

\begin{table}[ht]
\centering
\caption{Model sets used in the evaluated axes.}
\label{tab:model_axes_combined}
\resizebox{0.5\textwidth}{!}{
\begin{tabular}{l|l|l}
\toprule
\textbf{Qwen3 axis} & \textbf{Qwen2.5 axis} & \textbf{Cross-model axis} \\
\midrule
Qwen3-0.6B  & Qwen2.5-0.5B-Instruct   & Llama-3.2-1B-Instruct \\
Qwen3-1.7B  & Qwen2.5-1.5B-Instruct   & Llama-3.2-3B-Instruct \\
Qwen3-4B    & Qwen2.5-3B-Instruct     & Llama-3.1-8B-Instruct \\
Qwen3-8B    & Qwen2.5-7B-Instruct     & GLM-4-9B-0414 \\
Qwen3-14B   & Qwen2.5-14B-Instruct    & DeepSeek-R1-Distill-Qwen-14B \\
Qwen3-32B   & Qwen2.5-32B-Instruct    & GLM-4-32B-0414 \\
\bottomrule
\end{tabular}
}
\end{table}

\begin{table*}[ht]
\centering
\caption{Benchmark-level accuracy for the evaluated model sets across six benchmarks. The Qwen3 and Qwen2.5 axes follow benchmark-specific empirical orderings, while the heterogeneous cross-model setting uses benchmark-specific empirical orderings reported in Table~\ref{tab:cross_axis_task_order}.}
\label{tab:axis_accuracy_multi}
\resizebox{\textwidth}{!}{
\begin{tabular}{c|l|cccccc}
\toprule
\textbf{Axis} & \textbf{Model} & MMLU & GPQA & GSM8K & MATH & PopQA & TriviaQA \\
\midrule
\multirow{6}{*}{\textbf{\makecell{Qwen3\\Axis}}}
& Qwen3-0.6B & 41.20\% & 29.02\% & 68.16\% & 56.22\% & 12.94\% & 18.98\% \\
& Qwen3-1.7B & 61.15\% & 29.69\% & 85.29\% & 76.87\% & 15.08\% & 35.38\% \\
& Qwen3-4B   & 74.69\% & 37.28\% & 93.56\% & 86.00\% & 18.17\% & 52.20\% \\
& Qwen3-8B   & 78.97\% & 45.54\% & 94.39\% & 85.86\% & 23.38\% & 66.43\% \\
& Qwen3-14B  & 81.95\% & 50.89\% & 95.98\% & 88.12\% & 26.29\% & 71.95\% \\
& Qwen3-32B  & 85.22\% & 57.81\% & 95.98\% & 88.10\% & 28.31\% & 76.54\% \\
\midrule
\multirow{6}{*}{\textbf{\makecell{Qwen2.5\\Axis}}}
& Qwen2.5-0.5B-Instruct & 32.87\% & 25.22\% & 41.32\% & 24.88\% & 12.85\% & 26.07\% \\
& Qwen2.5-1.5B-Instruct & 57.85\% & 24.55\% & 66.26\% & 36.07\% & 15.83\% & 47.32\% \\
& Qwen2.5-3B-Instruct   & 69.36\% & 28.57\% & 90.67\% & 68.11\% & 17.94\% & 58.04\% \\
& Qwen2.5-7B-Instruct   & 74.21\% & 37.05\% & 94.01\% & 77.46\% & 21.31\% & 66.22\% \\
& Qwen2.5-14B-Instruct  & 80.89\% & 40.63\% & 95.22\% & 78.40\% & 26.31\% & 76.43\% \\
& Qwen2.5-32B-Instruct  & 83.71\% & 43.53\% & 95.91\% & 81.90\% & 26.39\% & 76.92\% \\
\midrule
\multirow{6}{*}{\textbf{\makecell{Cross-model\\Axis}}}
& Llama-3.2-1B-Instruct        & 48.65\% & 24.33\% & 40.56\% & 31.31\% & 17.42\% & 49.29\% \\
& Llama-3.2-3B-Instruct        & 65.51\% & 32.14\% & 86.81\% & 54.08\% & 24.19\% & 70.94\% \\
& Llama-3.1-8B-Instruct        & 75.51\% & 38.17\% & 92.04\% & 60.74\% & 34.34\% & 82.27\% \\
& GLM-4-9B-0414                & 81.76\% & 39.06\% & 93.40\% & 75.80\% & 23.76\% & 68.56\% \\
& DeepSeek-R1-Distill-Qwen-14B & 84.23\% & 35.49\% & 94.54\% & 81.67\% & 26.41\% & 77.81\% \\
& GLM-4-32B-0414               & 87.07\% & 53.57\% & 92.65\% & 89.28\% & 36.38\% & 84.49\% \\
\bottomrule
\end{tabular}
}
\end{table*}

\begin{table*}[ht]
\centering
\caption{Task-type-specific ordering of the heterogeneous cross-model axis. The same candidate model set is used, but the ordering is selected according to capability trends within each task type.}
\label{tab:cross_axis_task_order}
\resizebox{\textwidth}{!}{
\begin{tabular}{c|c|c}
\toprule
\textbf{Task type} & \textbf{Benchmarks} & \textbf{Ordered cross-model axis} \\
\midrule
MCQ 
& MMLU, GPQA 
& \makecell[l]{Llama-3.2-1B-Instruct $\rightarrow$ Llama-3.2-3B-Instruct $\rightarrow$ Llama-3.1-8B-Instruct $\rightarrow$\\
GLM-4-9B-0414 $\rightarrow$ DeepSeek-R1-Distill-Qwen-14B $\rightarrow$ GLM-4-32B-0414} \\
\midrule
Math 
& GSM8K, MATH 
& \makecell[l]{Llama-3.2-1B-Instruct $\rightarrow$ Llama-3.2-3B-Instruct $\rightarrow$ Llama-3.1-8B-Instruct $\rightarrow$\\
GLM-4-9B-0414 $\rightarrow$ DeepSeek-R1-Distill-Qwen-14B $\rightarrow$ GLM-4-32B-0414} \\
\midrule
QA 
& PopQA, TriviaQA 
& \makecell[l]{Llama-3.2-1B-Instruct $\rightarrow$ GLM-4-9B-0414 $\rightarrow$ Llama-3.2-3B-Instruct $\rightarrow$\\
DeepSeek-R1-Distill-Qwen-14B $\rightarrow$ Llama-3.1-8B-Instruct $\rightarrow$ GLM-4-32B-0414} \\
\bottomrule
\end{tabular}
}
\end{table*}

Table~\ref{tab:axis_accuracy_multi} reports benchmark-level accuracy for the evaluated model sets across six benchmarks. The Qwen3 and Qwen2.5 axes provide fixed within-family capability orderings, while the heterogeneous cross-model setting uses benchmark-specific empirical orderings shown in Table~\ref{tab:cross_axis_task_order}. This design avoids imposing a single global ordering across heterogeneous model families, whose relative strengths vary across MCQ, mathematical reasoning, and QA tasks. Local deviations are retained in the raw correctness trajectories and reflected by the order-consistency score $o_q$.

\section{Repeated-Response Reliability}
\label{app:observation_reliability}

The analysis reuses the original 50 Qwen3-8B responses per item across six benchmarks. For each item, responses are randomly partitioned into two disjoint blocks of 25, whose first $t$ responses form the comparison sets, with $t\in\{5,10,15,20,25\}$. State assignments and scoring outcomes remain fixed throughout.

Let $P_q^{A,t}$ and $P_q^{B,t}$ denote the empirical answer-state distributions of the two blocks for item $q$. Within-item discrepancy is $d_{q,t}^{\mathrm{native}}=\operatorname{JSD}(P_q^{A,t},P_q^{B,t})$, where
\begin{equation}
    \operatorname{JSD}(P,Q)
    =\frac{1}{2}\operatorname{KL}(P\Vert M)
    +\frac{1}{2}\operatorname{KL}(Q\Vert M),
\end{equation}
with $M=(P+Q)/2$. Natural logarithms give JSD a range of $[0,\log 2]$. This comparison uses the same item's answer states in both blocks.

\begin{table}[ht]
    \centering
    \small
    \setlength{\tabcolsep}{4pt}
    \caption{Mean within-item answer-state JSD between disjoint blocks of $t$ responses, using the first random partition and natural logarithms.}
    \label{tab:reliability_native_jsd}
    \begin{tabular}{lrrrrr}
        \toprule
        Benchmark & $t=5$ & $t=10$ & $t=15$ & $t=20$ & $t=25$ \\
        \midrule
        GPQA     & 0.0921 & 0.0504 & 0.0345 & 0.0261 & 0.0216 \\
        GSM8K    & 0.0182 & 0.0119 & 0.0101 & 0.0084 & 0.0073 \\
        MATH     & 0.0680 & 0.0513 & 0.0440 & 0.0394 & 0.0363 \\
        MMLU     & 0.0245 & 0.0145 & 0.0102 & 0.0080 & 0.0066 \\
        PopQA    & 0.1998 & 0.1442 & 0.1169 & 0.0998 & 0.0880 \\
        TriviaQA & 0.0593 & 0.0389 & 0.0299 & 0.0246 & 0.0214 \\
        \bottomrule
    \end{tabular}
\end{table}

\begin{table*}[ht]
    \centering
    \small
    \setlength{\tabcolsep}{6pt}
    \caption{Outcome-role JSD at $t=20$ and paired item-level bootstrap confidence intervals for $G_{20}=\text{Cross-JSD}-\text{Self-JSD}$. Item counts refer to the Qwen3-8B reliability cohort.}
    \label{tab:reliability_role_jsd}
    \begin{tabular}{lrrrrl}
        \toprule
        Benchmark & Items & Self-JSD & Cross-JSD & $G_{20}$ & 95\% CI for $G_{20}$ \\
        \midrule
        GPQA     & 448    & 0.0113 & 0.1878 & 0.1766 & [0.1631, 0.1893] \\
        GSM8K    & 1317  & 0.0025 & 0.0673 & 0.0648 & [0.0534, 0.0768] \\
        MATH     & 5000  & 0.0060 & 0.1521 & 0.1461 & [0.1394, 0.1529] \\
        MMLU     & 14042 & 0.0046 & 0.1909 & 0.1863 & [0.1822, 0.1904] \\
        PopQA    & 14267 & 0.0019 & 0.2184 & 0.2165 & [0.2115, 0.2212] \\
        TriviaQA & 7993  & 0.0024 & 0.2877 & 0.2853 & [0.2812, 0.2893] \\
        \bottomrule
    \end{tabular}
\end{table*}

Because answer states are item-specific, cross-item comparisons instead use the shared outcome roles $\mathcal U=$\{\texttt{correct}, \texttt{incorrect\_valid}, \texttt{invalid}\}, distinguishing correct answers, valid but incorrect answers, and outputs failing the recorded validity rules. Let $U_q^{A,t}$ and $U_q^{B,t}$ denote distributions over these roles. For a benchmark item set $\mathcal I$ of size $N$, within-item and cross-item discrepancies are
\begin{equation}
    d_{q,t}^{\mathrm{self}}
    =\operatorname{JSD}(U_q^{A,t},U_q^{B,t}),
\end{equation}
\begin{equation}
    d_{q,t}^{\mathrm{cross}}
    =\frac{1}{N-1}\sum_{q'\in\mathcal I\setminus\{q\}}
    \operatorname{JSD}(U_q^{A,t},U_{q'}^{B,t}).
\end{equation}
The contrast $G_t=N^{-1}\sum_{q\in\mathcal I}(d_{q,t}^{\mathrm{cross}}-d_{q,t}^{\mathrm{self}})$ compares their benchmark means, using all distinct ordered item pairs for the cross-item term.

At $t=20$, percentile 95\% confidence intervals for $G_{20}$ use 10000 item-level bootstrap replicates, keeping both response blocks paired and recomputing the contrast in each replicate. A one-sided permutation test uses 10000 permutations of block-$B$ item labels to assess whether the observed mean self-JSD is unusually small; Holm correction is applied across the six benchmark tests.

Mean within-item answer-state JSD decreases at every tested block-size increment on all six benchmarks, reaching 0.0080--0.0998 at $t=20$ (Table~\ref{tab:reliability_native_jsd}). Split-half Spearman correlations across items are at least 0.8466 for correctness mass. In the shared outcome-role space, cross-item JSD exceeds self-JSD by 0.0648--0.2853, with positive confidence intervals and Holm-adjusted $p<6\times10^{-4}$ on every benchmark (Table~\ref{tab:reliability_role_jsd}). These results support repeatable, item-dependent observations under the recorded protocol and state mapping. Profile sensitivity to response budgets is assessed in Appendix~\ref{app:axis_sensitivity}.

\section{Sampling and Axis Sensitivity}
\label{app:axis_sensitivity}

Sensitivity analyses reuse the stored responses from the main experiments, with the item cohort and empirical model ordering fixed within each benchmark--axis condition. The six-model setting with $T=20$ serves as the reference. Results below report the primary Qwen3 setting.

Response distributions and item profiles are recomputed for each condition, using order-score calibration matched to its model count and response budget (Appendix~\ref{app:profile_details}). Order-review agreement is the proportion of items receiving the same screening decision as in the reference setting, using $\widehat o_q^{\mathrm{ref}}\le0.05$. Boundary-position and boundary-width mean absolute errors (MAEs) are computed over items with positive fitted correctness gain in both conditions. These comparisons measure reproducibility relative to the main setting.

\subsection{Response-Budget Sensitivity}

For each of 100 repetitions, the 20 stored responses are independently permuted within each model--item pair. The first 5, 10, and 15 responses form nested subsets, while model ordering remains fixed. Table~\ref{tab:axis-budget} reports results averaged over repetitions. Larger budgets yield higher order-review agreement and smaller position and width errors across all six benchmarks.

\begin{table*}[ht]
    \centering
    \small
    \setlength{\tabcolsep}{4pt}
    \caption{Response-budget sensitivity on Qwen3 relative to $T=20$. Values are means over 100 subsampling repetitions. Agreement refers to item-level order-review decisions; position and width MAEs use items with positive fitted gain in both conditions.}
    \label{tab:axis-budget}
    \begin{tabular}{lrrrrrrrrr}
        \toprule
        & \multicolumn{3}{c}{$T=5$}
        & \multicolumn{3}{c}{$T=10$}
        & \multicolumn{3}{c}{$T=15$} \\
        \cmidrule(lr){2-4}
        \cmidrule(lr){5-7}
        \cmidrule(lr){8-10}
        Benchmark
        & Agree. (\%) & $x_q$ MAE & $w_q$ MAE
        & Agree. (\%) & $x_q$ MAE & $w_q$ MAE
        & Agree. (\%) & $x_q$ MAE & $w_q$ MAE \\
        \midrule
        GPQA
        & 64.8 & 0.080 & 0.096
        & 80.7 & 0.049 & 0.066
        & 89.8 & 0.029 & 0.041 \\
        GSM8K
        & 94.0 & 0.028 & 0.033
        & 96.7 & 0.016 & 0.021
        & 98.3 & 0.009 & 0.012 \\
        MATH
        & 93.6 & 0.051 & 0.064
        & 95.8 & 0.031 & 0.043
        & 97.6 & 0.018 & 0.026 \\
        MMLU
        & 82.2 & 0.032 & 0.041
        & 91.3 & 0.019 & 0.026
        & 95.9 & 0.011 & 0.016 \\
        PopQA
        & 91.0 & 0.027 & 0.032
        & 95.9 & 0.016 & 0.021
        & 98.2 & 0.009 & 0.013 \\
        TriviaQA
        & 87.7 & 0.019 & 0.023
        & 94.2 & 0.011 & 0.014
        & 97.4 & 0.006 & 0.009 \\
        \bottomrule
    \end{tabular}
\end{table*}

\subsection{Endpoint-Preserving Model Reduction}

At $T=20$, subaxes retain the first and last models and enumerate all intermediate-model subsets with $n=3$, $4$, or $5$, yielding four, six, and four subaxes, respectively. Retained models preserve their original ordering.

To compare boundary estimates on a common coordinate system, subaxis crossings are mapped piecewise linearly to the original six-model positions. A retained model with original index $i$ is assigned position $(i-1)/5$. Position error uses the mapped $Q_{0.50,q}$, and width error uses the difference between the mapped $Q_{0.75,q}$ and $Q_{0.25,q}$. Table~\ref{tab:axis-density} reports agreement ranges and mean errors across subaxes. Removing intermediate models can alter transition localization and order-review decisions even when both endpoints are retained.

\begin{table*}[ht]
    \centering
    \small
    \setlength{\tabcolsep}{4pt}
    \caption{Endpoint-preserving model reduction on Qwen3 relative to the six-model setting, with $T=20$. Agreement ranges cover all subaxes at each model count. Position and width MAEs are averaged across subaxes after mapping crossings to the original model coordinates.}
    \label{tab:axis-density}
    \begin{tabular}{lrrrrrrrrr}
        \toprule
        & \multicolumn{3}{c}{$n=3$}
        & \multicolumn{3}{c}{$n=4$}
        & \multicolumn{3}{c}{$n=5$} \\
        \cmidrule(lr){2-4}
        \cmidrule(lr){5-7}
        \cmidrule(lr){8-10}
        Benchmark
        & Agree. (\%) & $x_q$ MAE & $w_q$ MAE
        & Agree. (\%) & $x_q$ MAE & $w_q$ MAE
        & Agree. (\%) & $x_q$ MAE & $w_q$ MAE \\
        \midrule
        GPQA
        & 65.6--75.7 & 0.166 & 0.187
        & 75.0--86.4 & 0.114 & 0.133
        & 83.5--93.3 & 0.066 & 0.080 \\
        GSM8K
        & 93.9--94.8 & 0.142 & 0.161
        & 95.5--96.8 & 0.073 & 0.084
        & 97.3--98.2 & 0.033 & 0.038 \\
        MATH
        & 93.4--93.8 & 0.147 & 0.164
        & 95.1--95.5 & 0.080 & 0.095
        & 97.0--97.6 & 0.038 & 0.049 \\
        MMLU
        & 79.2--84.2 & 0.158 & 0.179
        & 85.1--90.5 & 0.095 & 0.109
        & 90.8--95.5 & 0.049 & 0.058 \\
        PopQA
        & 88.4--90.3 & 0.161 & 0.185
        & 92.1--94.5 & 0.102 & 0.112
        & 95.5--97.4 & 0.053 & 0.058 \\
        TriviaQA
        & 81.7--83.2 & 0.148 & 0.188
        & 87.5--89.0 & 0.091 & 0.112
        & 93.2--94.0 & 0.047 & 0.056 \\
        \bottomrule
    \end{tabular}
\end{table*}

\raggedbottom
\section{Representative Items}
\label{app:representative_cases}

The cards show one illustrative item per benchmark selected for review on Qwen3 ($\widehat o_q^{\mathrm{ref}}\le0.05$, $T=20$). Correct-response counts $\mathbf{k}_q$ follow the model order 0.6B, 1.7B, 4B, 8B, 14B, and 32B; for MATH, the order is 0.6B, 1.7B, 8B, 4B, 32B, and 14B. The minimum Monte Carlo score is $1/200{,}001\approx5.0\times10^{-6}$.

\begin{bsdcase}{MMLU}{logical\_fallacies\_6760}
\bsdcasefield{Question}{When someone tries to support a proposition with information that really has nothing to do with the claim being made, that person has probably committed which of the following fallacies?\\
A. straw person; B. genetic fallacy; C. irrelevant conclusion; D. ignoratio elenchi.}
\bsdcasefield{Reference}{C. irrelevant conclusion.}
\bsdcasefield{Responses}{All 120 responses select C or D. Qwen3-4B selects C throughout; Qwen3-1.7B and Qwen3-14B select D throughout and are scored incorrect.}
\bsdcasefield{Explanation}{C and D denote the same fallacy (\href{https://plato.stanford.edu/entries/fallacies/}{SEP: Fallacies}); the key scores equivalent answers differently.}
\bsdcasefooter{bsdoption}{Equivalent options}{20,0,20,19,0,1}{150.47}{5.0\times10^{-6}}
\end{bsdcase}

\begin{bsdcase}{GPQA}{gpqa\_main\_74}
\bsdcasefield{Question}{When 500 mL of PH$_3$ is decomposed the total volume of the reaction mixture becomes 600 mL only. The H$_2$ obtained in the above reaction is used to create electricity in a fuel cell. Calculate the volume of unreacted H$_2$ in the fuel cell when only 50 mL of O$_2$ is used.\\
A. 100 mL; B. 180 mL; C. 200 mL; D. 200 mL.}
\bsdcasefield{Reference}{200 mL; the source also lists 200 mL as an incorrect option.}
\bsdcasefield{Responses}{A Qwen3-14B response computes $300-100=200$ mL; all 20 final letters from this model are rejected by the recorded run key.}
\bsdcasefield{Explanation}{The correct-answer text occurs twice. A single accepted option cannot accommodate both copies. Displayed letters follow the processed item; screening uses run-specific option keys.}
\bsdcasefooter{bsdoption}{Duplicate options}{19,4,3,4,0,2}{61.30}{5.0\times10^{-6}}
\end{bsdcase}

\begin{bsdcase}{GSM8K}{gsm8k\_0952}
\bsdcasefield{Question}{Vince can staple 30 reports every 15 minutes. If he was stapling reports from 8:00 AM until 11:00 PM, how many reports did he staple altogether?}
\bsdcasefield{Reference}{360; the source solution uses a three-hour interval from 8 AM to 11 AM.}
\bsdcasefield{Responses}{Qwen3-1.7B returns 360 in 19/20 responses. Qwen3-8B, Qwen3-14B, and Qwen3-32B each return 1800 in 20/20 responses and are scored incorrect.}
\bsdcasefield{Explanation}{The interval is 15 hours, giving $15\times(60/15)\times30=1800$. The reference incorrectly assumes 11 AM.}
\bsdcasefooter{bsdreference}{Reference error}{6,19,3,0,0,0}{81.10}{5.0\times10^{-6}}
\end{bsdcase}

\begin{bsdcase}{MATH}{math\_prealgebra\_LLevel 4\_4060}
\bsdcasefield{Question}{Let $\boxed{N}$ mean the number of whole number divisors of $N$. For example, $\boxed{3}=2$, because 3 has two divisors, 1 and 3. Find the value of $\boxed{\boxed{11}\times\boxed{20}}$.}
\bsdcasefield{Reference}{The processed key is 12; the original solution explicitly gives 6.}
\bsdcasefield{Responses}{Qwen3-32B returns 6 in 20/20 responses and is scored incorrect; Qwen3-8B returns 12 throughout and is scored correct.}
\bsdcasefield{Explanation}{For the divisor-count operator $d$, $d(d(11)d(20))=d(12)=6$. Preprocessing extracts the last $\boxed{12}$, an operator argument, as the answer. The defect is in reference extraction; the original solution is correct.}
\bsdcasefooter{bsdpreprocessing}{Extraction error}{17,20,20,20,0,20}{90.95}{5.0\times10^{-6}}
\end{bsdcase}

\begin{bsdcase}{PopQA}{popqa\_6399207}
\bsdcasefield{Question}{What genre is My Family?}
\bsdcasefield{Reference}{Sitcom, referring to the British television series.}
\bsdcasefield{Responses}{Qwen3-8B returns British sitcom in 17/20 responses. Qwen3-1.7B returns drama in six responses and comedy-drama in twelve.}
\bsdcasefield{Explanation}{The title identifies both a \href{https://www.dltentertainment.com/Distribution/usa/168}{British sitcom} and the 1995 \href{https://catalog.afi.com/Catalog/moviedetails/60033}{drama film} \emph{My Family, Mi Familia}. The question omits the entity information needed to distinguish these referents.}
\bsdcasefooter{bsdambiguity}{Entity ambiguity}{0,0,0,17,0,18}{37.64}{1.0\times10^{-5}}
\end{bsdcase}

\begin{bsdcase}{TriviaQA}{triviaqa\_qf\_2576}
\bsdcasefield{Question}{Which African country is the worlds highest producer of cocoa?}
\bsdcasefield{Reference}{Ghana; C\^ote d'Ivoire and Ivory Coast are absent from its aliases.}
\bsdcasefield{Responses}{Qwen3-1.7B returns Ghana throughout. Qwen3-8B, Qwen3-14B, and Qwen3-32B return C\^ote d'Ivoire or Ivory Coast in 20/20 responses and are scored incorrect.}
\bsdcasefield{Explanation}{An \href{https://www.icco.org/wp-content/uploads/KeynoteThematic-2.-Philippe-Bastide.pdf}{ICCO-hosted 2017 presentation} (slide 5) places C\^ote d'Ivoire first in both 2006--2007 and 2016--2017. The question specifies no historical period supporting the Ghana reference.}
\bsdcasefooter{bsdreference}{Reference mismatch}{19,20,10,0,0,0}{126.63}{5.0\times10^{-6}}
\end{bsdcase}

\section{Reference-Axis Robustness of Subset Discriminability}
\label{app:cross_axis_transfer}
The main measurement-utility experiment reports NDR and Kendall $\tau_b$ using Qwen3 as the common reference axis, allowing subset-selection methods to be compared against the same evaluation models. To examine whether the observed discrimination gains and ranking agreement depend on this reference choice, the same BSDProbe-selected subsets are also evaluated on the Qwen2.5 and cross-model reference axes.

Model axes play two distinct roles in this analysis. The \textit{selection axis} is used by BSDProbe to construct item-level structural profiles and guide subset selection. The \textit{reference axis} specifies the models used to evaluate subset discrimination and ranking preservation. Within each reference-axis setting, baseline methods select subsets according to their own criteria and are evaluated using the same reference models as BSDProbe.

On the Qwen2.5 reference axis, Qwen3-selected BSDProbe subsets retain mean NDRs above one on all six benchmarks, showing discrimination gains beyond the selection models (Table~\ref{tab:ndr_qwen25_reference}). Using Qwen2.5 for selection further increases mean NDR on every benchmark, outperforming all five baselines on GPQA and PopQA with NDRs of 3.32 and 6.53, respectively. It closely matches TinyBenchmarks on MMLU and TriviaQA, although Bayesian achieves higher NDRs on GSM8K and MATH. Same-axis BSDProbe selection also achieves the highest or jointly highest mean Kendall $\tau_b$ on every benchmark, reaching 1.000 on five benchmarks and 0.970 on GPQA (Table~\ref{tab:measurement_kendall_tau_b_qwen25}). Its discrimination gains therefore accompany strong agreement with the full-benchmark model rankings.

On the cross-model reference axis, Qwen3-selected BSDProbe subsets again retain mean NDRs above one on all six benchmarks (Table~\ref{tab:ndr_cross_reference}). Cross-model selection achieves the highest mean NDR on MMLU, GPQA, and TriviaQA, reaching 2.12, 2.28, and 2.39, respectively. Bayesian has higher NDRs on GSM8K and MATH, while PSN-IRT leads on PopQA. Nevertheless, cross-model BSDProbe selection achieves the highest or jointly highest mean Kendall $\tau_b$ on every benchmark, reaching 1.000 on five benchmarks and 0.973 on GPQA (Table~\ref{tab:measurement_kendall_tau_b_cross_model}). The two metrics do not always favor the same selection axis: on GSM8K, Qwen3 selection yields a higher NDR than cross-model selection (1.43 versus 1.30), but lower ranking agreement (0.733 versus 1.000). Greater positive score separation can therefore coexist with weaker ranking preservation.

\begin{table*}[ht]
\centering
\caption{Subset discriminability measured by NDR for 100-item subsets evaluated on the Qwen2.5 reference axis ($\uparrow$ indicates higher discrimination).}
\label{tab:ndr_qwen25_reference}
\resizebox{\textwidth}{!}{
\setlength{\tabcolsep}{2pt}
\begin{tabular}{l|ccccc|ccc}
\toprule
\textbf{Benchmark} & Random & \makecell{Tiny\\Benchmarks} & \makecell{Anchor\\Points} & Bayesian & PSN-IRT & \makecell{BSDProbe\\(Qwen3)} & \makecell{BSDProbe\\(Qwen2.5)} & \makecell{BSDProbe\\(Cross-model)} \\
\midrule
MMLU     & $1.02_{\pm 0.12}$ & $\mathbf{1.57_{\pm 0.06}}$ & $0.89_{\pm 0.06}$ & $1.06_{\pm 0.15}$ & $1.12_{\pm 0.00}$ & $1.15_{\pm 0.06}$ & $1.57_{\pm 0.07}$ & $1.09_{\pm 0.09}$ \\
GPQA     & $1.13_{\pm 0.22}$ & $2.02_{\pm 0.22}$ & $1.59_{\pm 0.31}$ & $2.02_{\pm 0.17}$ & $2.53_{\pm 0.00}$ & $1.88_{\pm 0.12}$ & $\mathbf{3.32_{\pm 0.16}}$ & $2.14_{\pm 0.20}$ \\
\midrule
GSM8K    & $1.07_{\pm 0.07}$ & $1.31_{\pm 0.06}$ & $0.96_{\pm 0.08}$ & $\mathbf{1.55_{\pm 0.04}}$ & $1.42_{\pm 0.00}$ & $1.34_{\pm 0.05}$ & $1.38_{\pm 0.07}$ & $1.26_{\pm 0.07}$ \\
MATH     & $0.99_{\pm 0.06}$ & $1.47_{\pm 0.03}$ & $1.22_{\pm 0.03}$ & $\mathbf{1.56_{\pm 0.06}}$ & $0.84_{\pm 0.00}$ & $1.02_{\pm 0.07}$ & $1.30_{\pm 0.06}$ & $1.21_{\pm 0.09}$ \\
\midrule
PopQA    & $0.98_{\pm 0.27}$ & $0.10_{\pm 0.05}$ & $1.93_{\pm 0.22}$ & $3.18_{\pm 0.40}$ & $5.61_{\pm 0.00}$ & $4.42_{\pm 0.31}$ & $\mathbf{6.53_{\pm 0.25}}$ & $3.34_{\pm 0.39}$ \\
TriviaQA & $1.03_{\pm 0.09}$ & $\mathbf{1.78_{\pm 0.06}}$ & $0.85_{\pm 0.04}$ & $1.36_{\pm 0.10}$ & $1.57_{\pm 0.00}$ & $1.40_{\pm 0.07}$ & $1.78_{\pm 0.05}$ & $1.17_{\pm 0.07}$ \\
\bottomrule
\end{tabular}
}
\end{table*}

\begin{table*}[ht]
\centering
\caption{Ranking agreement measured by Kendall $\tau_b$ for 100-item subsets evaluated on the Qwen2.5 reference axis ($\uparrow$ indicates higher agreement).}
\label{tab:measurement_kendall_tau_b_qwen25}
\resizebox{\textwidth}{!}{
\setlength{\tabcolsep}{2pt}
\begin{tabular}{l|ccccc|ccc}
\toprule
\textbf{Benchmark} & Random & \makecell{Tiny\\Benchmarks} & \makecell{Anchor\\Points} & Bayesian & PSN-IRT & \makecell{BSDProbe\\(Qwen3)} & \makecell{BSDProbe\\(Qwen2.5)} & \makecell{BSDProbe\\(Cross-model)} \\
\midrule
MMLU     & $0.980_{\pm 0.042}$ & $0.949_{\pm 0.044}$ & $0.997_{\pm 0.011}$ & $0.913_{\pm 0.061}$ & $\mathbf{1.000_{\pm 0.000}}$ & $0.997_{\pm 0.011}$ & $\mathbf{1.000_{\pm 0.000}}$ & $0.997_{\pm 0.011}$ \\
GPQA     & $0.794_{\pm 0.152}$ & $0.848_{\pm 0.124}$ & $0.765_{\pm 0.063}$ & $0.900_{\pm 0.054}$ & $0.867_{\pm 0.000}$ & $0.842_{\pm 0.095}$ & $\mathbf{0.970_{\pm 0.055}}$ & $0.877_{\pm 0.031}$ \\
\midrule
GSM8K    & $0.824_{\pm 0.157}$ & $0.915_{\pm 0.060}$ & $0.896_{\pm 0.079}$ & $0.987_{\pm 0.042}$ & $0.894_{\pm 0.000}$ & $0.997_{\pm 0.011}$ & $\mathbf{1.000_{\pm 0.000}}$ & $0.896_{\pm 0.048}$ \\
MATH     & $0.963_{\pm 0.063}$ & $0.859_{\pm 0.016}$ & $0.873_{\pm 0.035}$ & $0.976_{\pm 0.054}$ & $0.966_{\pm 0.000}$ & $0.960_{\pm 0.064}$ & $\mathbf{1.000_{\pm 0.000}}$ & $0.959_{\pm 0.059}$ \\
\midrule
PopQA    & $0.802_{\pm 0.161}$ & $-0.438_{\pm 0.299}$ & $0.960_{\pm 0.052}$ & $0.903_{\pm 0.093}$ & $\mathbf{1.000_{\pm 0.000}}$ & $0.930_{\pm 0.067}$ & $\mathbf{1.000_{\pm 0.000}}$ & $0.970_{\pm 0.055}$ \\
TriviaQA & $0.926_{\pm 0.096}$ & $0.963_{\pm 0.053}$ & $0.909_{\pm 0.065}$ & $0.927_{\pm 0.090}$ & $\mathbf{1.000_{\pm 0.000}}$ & $0.983_{\pm 0.042}$ & $\mathbf{1.000_{\pm 0.000}}$ & $0.973_{\pm 0.056}$ \\
\bottomrule
\end{tabular}
}
\end{table*}

\begin{table*}[ht]
\centering
\caption{Subset discriminability measured by NDR for 100-item subsets evaluated on the cross-model reference axis ($\uparrow$ indicates higher discrimination).}
\label{tab:ndr_cross_reference}
\resizebox{\textwidth}{!}{
\setlength{\tabcolsep}{2pt}
\begin{tabular}{l|ccccc|ccc}
\toprule
\textbf{Benchmark} & Random & \makecell{Tiny\\Benchmarks} & \makecell{Anchor\\Points} & Bayesian & PSN-IRT & \makecell{BSDProbe\\(Qwen3)} & \makecell{BSDProbe\\(Qwen2.5)} & \makecell{BSDProbe\\(Cross-model)} \\
\midrule
MMLU     & $1.10_{\pm 0.10}$ & $1.08_{\pm 0.09}$ & $0.88_{\pm 0.06}$ & $1.31_{\pm 0.19}$ & $1.72_{\pm 0.25}$ & $1.60_{\pm 0.15}$ & $1.41_{\pm 0.14}$ & $\mathbf{2.12_{\pm 0.13}}$ \\
GPQA     & $1.07_{\pm 0.19}$ & $1.80_{\pm 0.17}$ & $1.06_{\pm 0.11}$ & $1.62_{\pm 0.23}$ & $1.92_{\pm 0.00}$ & $1.88_{\pm 0.11}$ & $1.61_{\pm 0.18}$ & $\mathbf{2.28_{\pm 0.10}}$ \\
\midrule
GSM8K    & $1.01_{\pm 0.12}$ & $1.48_{\pm 0.09}$ & $1.11_{\pm 0.07}$ & $\mathbf{1.53_{\pm 0.15}}$ & $1.46_{\pm 0.00}$ & $1.43_{\pm 0.11}$ & $1.31_{\pm 0.07}$ & $1.30_{\pm 0.06}$ \\
MATH     & $0.95_{\pm 0.07}$ & $1.31_{\pm 0.03}$ & $0.97_{\pm 0.04}$ & $\mathbf{1.54_{\pm 0.04}}$ & $0.81_{\pm 0.00}$ & $1.29_{\pm 0.06}$ & $1.20_{\pm 0.09}$ & $1.30_{\pm 0.05}$ \\
\midrule
PopQA    & $1.04_{\pm 0.21}$ & $0.65_{\pm 0.12}$ & $2.17_{\pm 0.25}$ & $3.13_{\pm 0.43}$ & $\mathbf{4.59_{\pm 0.00}}$ & $2.66_{\pm 0.25}$ & $2.46_{\pm 0.13}$ & $4.41_{\pm 0.26}$ \\
TriviaQA & $1.04_{\pm 0.11}$ & $0.99_{\pm 0.11}$ & $1.11_{\pm 0.08}$ & $1.69_{\pm 0.14}$ & $1.80_{\pm 0.16}$ & $1.48_{\pm 0.07}$ & $1.59_{\pm 0.15}$ & $\mathbf{2.39_{\pm 0.11}}$ \\
\bottomrule
\end{tabular}
}
\end{table*}

\begin{table*}[ht]
\centering
\caption{Ranking agreement measured by Kendall $\tau_b$ for 100-item subsets evaluated on the cross-model reference axis ($\uparrow$ indicates higher agreement).}
\label{tab:measurement_kendall_tau_b_cross_model}
\resizebox{\textwidth}{!}{
\setlength{\tabcolsep}{2pt}
\begin{tabular}{l|ccccc|ccc}
\toprule
\textbf{Benchmark} & Random & \makecell{Tiny\\Benchmarks} & \makecell{Anchor\\Points} & Bayesian & PSN-IRT & \makecell{BSDProbe\\(Qwen3)} & \makecell{BSDProbe\\(Qwen2.5)} & \makecell{BSDProbe\\(Cross-model)} \\
\midrule
MMLU     & $0.926_{\pm 0.096}$ & $0.927_{\pm 0.036}$ & $0.997_{\pm 0.011}$ & $0.899_{\pm 0.115}$ & $0.920_{\pm 0.069}$ & $0.987_{\pm 0.042}$ & $0.960_{\pm 0.090}$ & $\mathbf{1.000_{\pm 0.000}}$ \\
GPQA     & $0.862_{\pm 0.113}$ & $0.957_{\pm 0.063}$ & $0.742_{\pm 0.095}$ & $0.837_{\pm 0.118}$ & $0.733_{\pm 0.000}$ & $0.859_{\pm 0.056}$ & $0.963_{\pm 0.053}$ & $\mathbf{0.973_{\pm 0.041}}$ \\
\midrule
GSM8K    & $0.720_{\pm 0.134}$ & $0.450_{\pm 0.245}$ & $0.695_{\pm 0.091}$ & $0.743_{\pm 0.030}$ & $0.733_{\pm 0.000}$ & $0.733_{\pm 0.000}$ & $0.765_{\pm 0.063}$ & $\mathbf{1.000_{\pm 0.000}}$ \\
MATH     & $0.983_{\pm 0.042}$ & $\mathbf{1.000_{\pm 0.000}}$ & $0.970_{\pm 0.055}$ & $\mathbf{1.000_{\pm 0.000}}$ & $0.966_{\pm 0.000}$ & $\mathbf{1.000_{\pm 0.000}}$ & $0.987_{\pm 0.042}$ & $\mathbf{1.000_{\pm 0.000}}$ \\
\midrule
PopQA    & $0.912_{\pm 0.066}$ & $0.703_{\pm 0.132}$ & $0.747_{\pm 0.076}$ & $0.815_{\pm 0.125}$ & $0.733_{\pm 0.000}$ & $0.886_{\pm 0.104}$ & $0.922_{\pm 0.099}$ & $\mathbf{1.000_{\pm 0.000}}$ \\
TriviaQA & $0.884_{\pm 0.069}$ & $0.454_{\pm 0.267}$ & $0.903_{\pm 0.068}$ & $0.997_{\pm 0.011}$ & $0.969_{\pm 0.065}$ & $0.963_{\pm 0.063}$ & $0.896_{\pm 0.067}$ & $\mathbf{1.000_{\pm 0.000}}$ \\
\bottomrule
\end{tabular}
}
\end{table*}

\end{document}